\documentclass{article}

\usepackage[final]{neurips_2022}

\usepackage[medium,compact]{titlesec}

\usepackage{setspace}
\usepackage{diagbox}
\usepackage[table,xcdraw]{xcolor}
\usepackage[utf8]{inputenc} 
\usepackage[T1]{fontenc}    
\usepackage{url}            
\usepackage{booktabs}       
\usepackage{amsfonts}       
\usepackage{nicefrac}       
\usepackage{microtype}      
\usepackage{xcolor}         
\usepackage{times}
\usepackage{epsfig}
\usepackage{graphicx}
\usepackage{amsmath}
\usepackage{amssymb}
\usepackage{dsfont}
\usepackage{multirow}
\usepackage{adjustbox}
\usepackage{wrapfig}
\newcommand{\sign}{\text{sign}}

\newtheorem{myAss}{Assumption}
\newtheorem{myTheo}{Theorem}

\newtheorem{proof}{Proof}
\usepackage[ruled,linesnumbered]{algorithm2e}
\usepackage{subfigure}
\makeatletter

\newcommand{\Rmnum}[1]{\expandafter\@slowromancap\romannumeral #1@}

\newcommand{\etal}{\emph{et~al.}\xspace}
\newcommand{\eg}{\emph{e.g.},\xspace}
\newcommand{\ie}{\emph{i.e.},\xspace}

\newcommand{\etc}{\emph{etc.}\xspace}

\DeclareMathAlphabet\mathbfcal{OMS}{cmsy}{b}{n}

\newcommand{\eat}[1]{}
\ifodd 1
\newcommand{\rev}[1]{{\color{purple}{#1}}} 

\newcommand{\fan}[1]{{\color{brown}{#1}}}

\newcommand{\TODO}[1]{{\color{red}TODO:{#1}}}

\else
\newcommand{\rev}[1]{#1}

\newcommand{\TODO}[1]{}

\fi

\title{Practical Adversarial Attacks on Spatiotemporal Traffic Forecasting Models}

\author{%
  Fan~LIU \\
  AI Thrust\&RBM, The Hong Kong University of Science and Technology (Guangzhou) \\
  \texttt{fliu236@connect.hkust-gz.edu.cn\&liufan@ust.hk} \\
   \And
  Hao~LIU\thanks{Corresponding author} \\
  AI Thrust, The Hong Kong University of Science and Technology (Guangzhou) \\ Guangzhou HKUST Fok Ying Tung Research Institute \\ CSE, The Hong Kong University of Science and Technology\\
  \texttt{liuh@ust.hk} \\
   \AND
    Wenzhao~Jiang \\
  AI Thrust, The Hong Kong University of Science and Technology (Guangzhou) \\
  \texttt{wjiang431@connect.hkust-gz.edu.cn} \\
}

\begin{document}

\maketitle

\begin{abstract}

Machine learning based traffic forecasting models leverage sophisticated spatiotemporal auto-correlations to provide accurate predictions of city-wide traffic states.
However, existing methods assume a reliable and unbiased forecasting environment, which is not always available in the wild. In this work, we investigate the vulnerability of spatiotemporal traffic forecasting models and propose a practical adversarial spatiotemporal attack framework.
Specifically, instead of simultaneously attacking all geo-distributed data sources, an iterative gradient-guided node saliency method is proposed to identify the time-dependent set of victim nodes.
Furthermore, we devise a spatiotemporal gradient descent based scheme to generate real-valued adversarial traffic states under a perturbation constraint.
Meanwhile, we theoretically demonstrate the worst performance bound of adversarial traffic forecasting attacks.
Extensive experiments on two real-world datasets show that the proposed two-step framework achieves up to $67.8\%$ performance degradation on various advanced spatiotemporal forecasting models.
Remarkably, we also show that adversarial training with our proposed attacks can significantly improve the robustness of spatiotemporal traffic forecasting models.
Our code is available in \url{https://github.com/kdd-hkust/Adv-ST}.

\end{abstract}
\section{Introduction}

\begin{figure}[t]
    \centering
\includegraphics[width=0.9\textwidth]{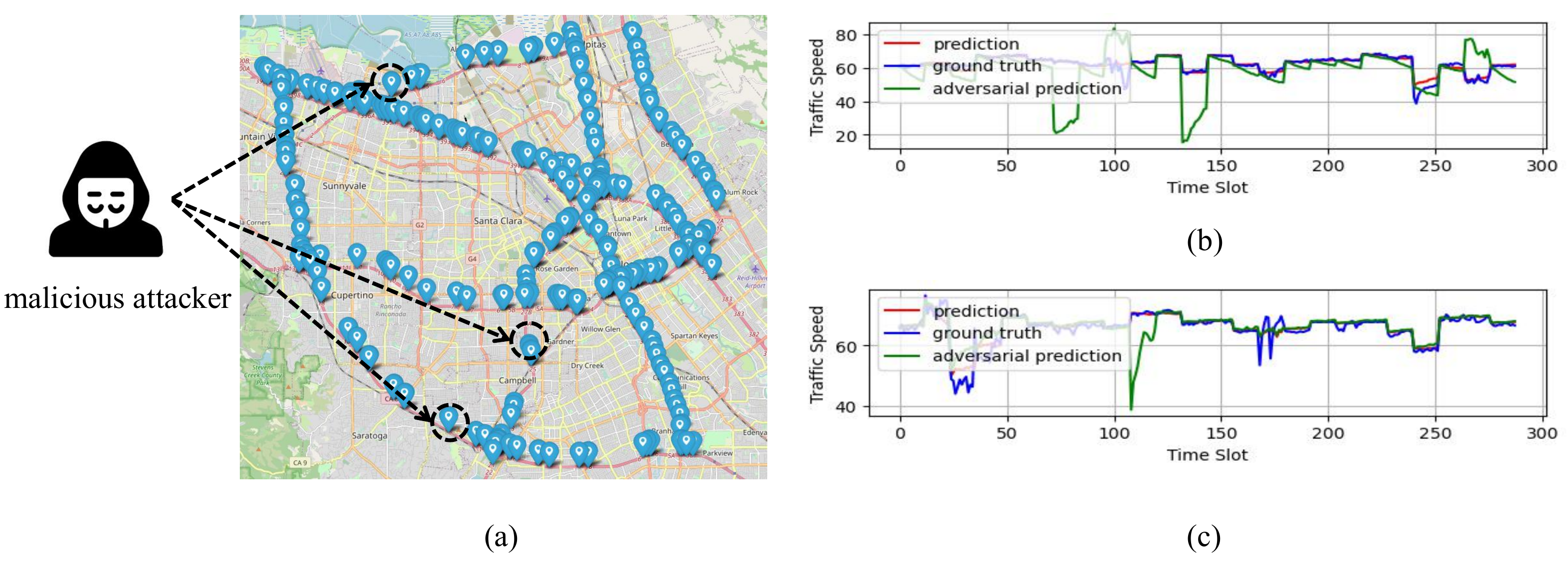}
\caption{An illustration of adversarial attack against spatiotemporal forecasting models on the Bay Area traffic network in California, the data ranges from January 2017 to May 2017.
(a) Adversarial attack of geo-distributed data.  The malicious attacker may inject adversarial examples into a few randomly selected geo-distributed data sources.~(\eg roadway sensors) to mislead the prediction of the whole traffic forecasting system.
(b) Accuracy drop of victim nodes. By adding less than 50\% traffic speed perturbations to 10\% victim nodes, we observe 60.4\% accuracy drop of victim nodes in  morning peak hour.
(c) Accuracy drop of neighbouring nodes. Due to the information diffusion of spatiotemporal forecasting models,
the adversarial attack also leads to up to about 47.23\% accuracy drop for neighboring nodes.
}\label{fig:content}
\end{figure}

Machine learned spatiotemporal forecasting models have been widely adopted in modern Intelligent Transportation Systems (ITS) to provide accurate and timely prediction of traffic dynamics, \eg traffic flow~\cite{chen2021trafficstream}, traffic speed~\cite{Graph_Wave_Net, LiaoZWMCYGW18}, and the estimated time of arrival~\cite{qiu2019nei, derrow2021eta}.
Despite fruitful progress in improving the forecasting accuracy and utility~\cite{AGCRN}, little attention has been paid to the robustness of spatiotemporal forecasting models.
\eat{For example, Fig. \rev{Use Figure, don't use abbreviation.} \ref{fig:content} shows that
carefully designed  perturbations on a few nodes induce significant performance degradation of the whole forecasting system.
In this paper, we investigate the adversarial robustness of traffic forecasting models against adversarial attacks in the wild.}
For example, Figure~\ref{fig:content} demonstrates that injecting slight adversarial perturbations on a few randomly selected nodes can significantly degrade the traffic forecasting accuracy of the whole system.
Therefore, this paper investigates the vulnerability of traffic forecasting models against adversarial attacks.

In recent years, adversarial attacks have been extensively studied in various application domains, such as computer vision and natural language processing~\cite{xu2020adversarial}
However, two major challenges prevent applying existing adversarial attack strategies to spatiotemporal traffic forecasting.
First, the traffic forecasting system makes predictions by exploiting signals from geo-distributed data sources~(\eg hundreds of roadway sensors and thousands of in-vehicle GPS devices). It is expensive and impractical to manipulate all data sources to inject adversarial perturbations simultaneously.
Furthermore, state-of-the-art traffic forecasting models propagate local traffic states through the traffic network for more accurate prediction~\cite{derrow2021eta}. Attacking a few arbitrary data sources will result in node-varying effects on the whole system.
How to identify the subset of salient victim nodes with a limited attack budget to maximize the attack effect is the first challenge.
Second, unlike most existing adversarial attack strategies that focus on time-invariant label classification~\cite{madry2018towards-PGD, dong2018boosting-MIM}, the adversarial attack against traffic forecasting aims to disrupt the target model to make biased predictions of continuous traffic states.
How to generate real-valued adversarial examples without access to the ground truth of future traffic states is another challenge.

\eat{To this end, in this paper, we propose a two-step framework to attack spatiotemporal traffic forecasting models by generating time-dependent adversarial traffic states. We show that attacking $10\%$ nodes can break down the forecasting accuracy from 1.975 to 6.1329, as depicted in Figure \ref{fig:content} (b-c).}
 \eat{Specifically, our framework consists of two steps: identifying the time-dependent victim nodes with a limited number budget and disturbing the nodes (sensors) through the adversarial spatiotemporal feature with maximum bound perturbation.} \eat{We proposed a time-dependent node saliency technique to  identify the victim nodes.}  \eat{Furthermore, we employ a surrogate model to estimate  future traffic conditions based on previous traffic states for time-dependent data. By design, we randomly sample the noise from a pre-defined distribution to finally guide the attack direction.}

 To this end, in this paper, we propose a practical adversarial spatiotemporal attack framework that can disrupt the forecasting models to derive biased city-wide traffic predictions.
 Specifically, we first devise an iterative gradient-guided method to estimate node saliency, which helps to identify a small time-dependent set of victim nodes.
 Moreover, a spatiotemporal gradient descent scheme is proposed to guide the attack direction and generate real-valued adversarial traffic states under a human imperceptible perturbation constraint.
 The proposed attack framework is agnostic to forecasting model architecture and is generalizable to various attack settings, \ie white-box attack, grey-box attack, and black-box attack.
 Meanwhile, we theoretically analyze the worst performance guarantees of adversarial traffic forecasting attacks. We prove the adversarial robustness of spatiotemporal traffic forecasting models is related to the number of victim nodes, the maximum perturbation bound, and the maximum degree of the traffic network.

 Extensive experimental studies on two real-world traffic datasets demonstrate the attack effectiveness of the proposed framework on state-of-the-art spatiotemporal forecasting models.
 We show that attacking $10\%$ nodes in the traffic system can break down the global forecasting Mean Average Error~(MAE) from $1.975$ to $6.1329$.
 Moreover, the adversarial attack can induce $68.65\%$, and $56.67\%$ performance degradation under the extended white-box and black-box attack settings, respectively.
 Finally, we also show that incorporating adversarial examples we generated with adversarial training can significantly improve the robustness of spatiotemporal traffic forecasting models.

 \section{Background and problem statement}
In this section, we first introduce some basics of spatiotemporal traffic forecasting and adversarial attack, then formally define the problem we aim to address.
\subsection{Spatiotemporal traffic forecasting}

Let $\mathcal{G}_{t}= (\mathcal{V}, \mathcal{E})$ denote a traffic network at time step $t$, where $\mathcal{V}$ is a set of $n$ nodes~(\eg regions, road segments, roadway sensors, \etc) and $\mathcal{E}$ is a set of edges. The construction of $\mathcal{G}_{t}$ can be categorized into two types, (1) prior-based, which pre-define $\mathcal{G}_{t}$ based on metrics such as geographical proximity and similarity~\cite{STGCN}, and (2) learning-based, which automatically learns $\mathcal{G}_{t}$ in an end-to-end way~\cite{Graph_Wave_Net}.
Note the $\mathcal{G}_{t}$ can be static or time-evolving depending on the forecasting model.
We denote $\mathbf{X}_{t}= (\mathbf{x}_{1,t}, \mathbf{x}_{2,t}, \cdots, \mathbf{x}_{n,t})$ as the spatiotemporal features associated to $\mathcal{G}_{t}$, where $\mathbf{x}_{i,t}\in\mathbb{R}^{c}$ represents the $c$-dimensional time-varying traffic conditions (\eg traffic volume, traffic speed) and contextual features~(\eg weather, surrounding POIs)
of node $v_i\in \mathcal{V}$ at $t$.
The spatiotemporal traffic forecasting problem aims to predict traffic states for all $v_i\in \mathcal{V}$ over the next $\tau$ time steps,
\begin{equation}
 \mathbf{\hat{Y}}_{t+1:t+\tau}=f_{\theta }(\mathbfcal{H}_{t-T+1:t}),
\end{equation}

where $\mathbfcal{H}_{t-T+1:t}=\{(\mathbf{X}_{t-T+1},\mathcal{G}_{t-T+1}),\dots, (\mathbf{X}_{t},\mathcal{G}_{t})\}$ denotes the traffic states contains input features and the traffic network in previous $T$ time steps, $f_{\theta }(\cdot)$ is the spatiotemporal traffic forecasting model parameterized by $\theta$, and $\mathbf{\hat{Y}}_{t+1:t+\tau} = \{ \mathbf{\hat{Y}}_{t+1},\mathbf{\hat{Y}}_{t+2},\cdots, \mathbf{\hat{Y}}_{t+\tau}\}$ is the estimated traffic conditions of interest of $\mathcal{V}$ from time step $t+1$ to $t+\tau$. We denote $ \mathbf{Y}_{t+1:t+\tau} = \{ \mathbf{Y}_{t+1},\mathbf{Y}_{t+2},\cdots, \mathbf{Y}_{t+\tau}\}$ as the ground truth of $\mathbfcal{H}_{t-T+1:t}.$

Note the above formulation is consistent with the state-of-the-art Graph Neural Network~(GNN) based spatiotemporal traffic forecasting models~\cite{Graph_Wave_Net, STGCN, ASTGCN, STAWnet}, and is also generalizable to other variants such as Convolutional Neural Network~(CNN) based approaches~\cite{zhang2017deep}.

\eat{
Let the $k$-th layer embedding of spatiotemporal traffic models is  $\mathbf{z}_{i}^{(k)} = \mathbf{h}_{i}^{(k)}\mathbf{W}^{(k)}$, where $\mathbf{h}_{i}^{(k+1)}  = \sigma (\sum_{j\in \mathcal{N}_{i}}e_{ij}\mathbf{z}_{j}^{(k)})$, and  $\mathbf{h}_{i}^{(k)}$ ($\mathbf{h^{(0)}}=\mathbfcal{H}_{t-T+1:t}$) is the $L$ layer spatiotemporal forecasting model's $k$-th layer embedding for node $i$     and $\mathbf{W}^{(k)}$ is the weight matrix.  $\sigma$ is an activation function, such as the sigmoid function, relu function, and so on. $e_{ij}$ is the weight value used to aggregate node $j$'s neighbors. $\mathcal{N}_{i}$ is the index used to keep track of node $j$'s neighbors. }

\subsection{Adversarial attack}
Given a machine learning model, adversarial attack aims to mislead the model to derive biased predictions by generating the optimal adversarial example
\begin{equation}
       x^{\ast }\in \arg \max_{x^\prime}\mathcal{L}(x^\prime,y;\theta ) \quad s.t. \left \| x^\prime-x \right \|_{p}\le  \varepsilon,
\end{equation}
where $x^\prime$ is the adversarial example with maximum bound $\varepsilon$ under $L_p$ norm to guarantee the perturbation is imperceptible to human, and $y$ is the ground truth of clean example $x$.

Various gradient-based methods have been proposed to generate adversarial examples, such as FGSM~\cite{goodfellow2014explaining}, PGD~\cite{madry2018towards-PGD}, MIM~\cite{dong2018boosting-MIM}, \etc
For instance, the adversarial example $x^\prime = x +\varepsilon\sign (\nabla_{x}\mathcal{L}_{CE}(x,y;\theta))$ in FGSM, where $\sign(\cdot)$ is the Signum function and $\mathcal{L}_{CE}(\cdot)$ is the cross entropy loss.

Note the adversarial attack happened in the testing stage, and the attackers cannot manipulate the forecasting model or its output. On the benign testing set, the forecasting model can perform well. Based on the amount of information the attacker can access in the testing stage, the adversarial attack can be categorized into three classes.
\emph{White-box attack}. The attacker can fully access the target model, including the model architecture, the model parameters, gradients, model outputs, the input traffic states, and the corresponding labels.
\emph{Grey-box attack}. The attacker can partially access the system, including the target model and the input traffic states, but without the labels.
\emph{Black-box attack}. The attacker can only access the input traffic states,  query the outputs of the target model or leverage a surrogate model to craft the adversarial examples.

\subsection{Adversarial attack against spatiotemporal traffic forecasting}
This work aims to apply adversarial attacks to spatiotemporal traffic forecasting models.
We first define the \emph{adversarial traffic state} as follow,
\begin{equation}
      \mathbfcal{H}^{\prime}_{t} = \left \{(\mathbf{X}^{\prime}_{t}, \mathcal{G}_{t}):   \left \|  S_{t} \right \|_{0} \le \eta , \left \| (\mathbf{X}^{\prime}_{t} -\mathbf{X}_{t})\cdot S_{t}  \right \|_{p} \le \varepsilon \right\},
\end{equation}
where $S_{t} \in \{ 0,1 \}^{n \times n}$ is a diagonal matrix with $i$th diagonal element indicating whether node $i$ is a victim node, and $\mathbf{X}^{\prime}_{t}$ is the perturbed spatiotemporal feature named adversarial spatiotemporal feature.
We restrict the adversarial traffic state by the victim node budget $\eta$ and the perturbation budget $\varepsilon$.

Note that following the definition of adversarial attack, we leave the topology of $\mathcal{G}_{t}$ immutable as we regard the adjacency relationship as a part of the model parameter that may be automatically learned in an end-to-end way.

\emph{Attack goal}. The attacker aims to craft adversarial traffic states to fool the spatiotemporal forecasting model to derive biased predictions. Formally, given a spatiotemporal forecasting model $f_\theta(\cdot)$, the adversarial attack against spatiotemporal traffic forecasting is defined as
\begin{subequations}\label{eq:ST_adversarial_attack}
\begin{align}
    &\max_{\substack{\mathbfcal{H}^{\prime}_{t-T+1:t} \\ t \in \mathcal{T}_{test}}}\sum_{t\in \mathcal{T}_{test}} \mathcal{L}(f_{\theta^{\ast } }(\mathbfcal{H}^{\prime}_{t-T+1:t} ), \mathbf{Y}_{t+1:t+\tau})\\
    &s.t., \quad \theta^{\ast }= \arg \min_{\theta }\sum_{t \in \mathcal{T}_{train}}\mathcal{L}(f_{\theta }(\mathbfcal{H}_{t-T+1:t}),\mathbf{Y}_{t+1:t+\tau})\label{eq:optimal_traing},
\end{align}\normalsize
\end{subequations}

where $\mathcal{T}_{test}$  and $\mathcal{T}_{train}$ denote the set of time steps of all testing and training samples, respectively. $\mathcal{L}(\cdot)$ is the loss function measuring the distance between the predicted traffic states and ground truth, and $\theta^{\ast}$ is optimal parameters learned during the training stage.

Since the ground truth~(\ie future traffic states) under the spatiotemporal traffic forecasting setting is unavailable at run-time, the practical adversarial spatiotemporal attack primarily falls into the grey-box attack setting.

However, investigating white-box attacks is still beneficial to help us understand how adversarial attack works and can help improve the robustness of spatiotemporal traffic forecasting models~(\eg apply adversarial training).
We discuss how to extend our proposed adversarial attack framework to white-box and black-box settings in Section~\ref{sec:attack-traffic}.

\section{Methodology}\label{sec:framework}
In this section, we introduce the practical adversarial spatiotemporal attack framework in detail. Specifically,  our framework consists of two steps: (1)~identify the  time-dependent victim nodes, and (2)~attack with the adversarial traffic state.

\subsection{Identify time-dependent victim nodes}
\eat{ Since the attacker needs present traffic states $\mathbfcal{H}_{t}$ to identify the victim nodes  in advance, we use the traffic state prediction function to obtain the $\mathbfcal{H}_{t}$ based on past traffic state $\mathbfcal{H}_{t-1}$.   The traffic state prediction function is defined as,
\begin{equation}\label{eq:pseudo_traffic_state}
   g_{\varphi}(\mathbfcal{H}_{t-1}) = \mathbfcal{\tilde{H}}_{t},
\end{equation}
where $g_{\varphi}(\cdot)$ is the function to estimate the traffic states. We have $\mathbfcal{\tilde{H}}_{t} \approx \mathbfcal{H}_{t}$.  The estimated  traffic states $\mathbf{\tilde{H}}_{t}$ can help the attacker  prepare to identify at time step $t$.  In our implementation,  we can obtain $g_{\varphi}(\cdot)$ by pre-training.}

One unique characteristic that distinguishes attacking spatiotemporal forecasting from conventional classification tasks is the inaccessibility of ground truth at the test phase.
Therefore, we first construct future traffic states' surrogate label to guide the attack direction,
\begin{equation}\label{eq:pseudo_label}
 \mathbf{\tilde{Y}}_{t+1:t+\tau} =  g_{\phi}(\mathbfcal{H}_{t-T+1:t}) + \mathcal{\delta}_{t+1:t+\tau},
\end{equation}
where $g_{\phi}(\cdot)$ is a generalized function~(\eg $\tanh(\cdot)$, $\sin{(\cdot)}$, $f_{\theta}(\cdot))$,
$\mathcal{\delta}_{t+1:t+\tau}$ are random variables sampled from a probability distribution $\pi(\mathcal{\delta}_{t+1:t+\tau})$ to increase the diversity of the attack direction. In our implementation, we derive $\phi$ based on the pre-trained forecasting model parameter $\theta^{\ast}$, and $\mathcal{\delta}_{t+1:t+\tau} \sim U(-\varepsilon /10, \varepsilon /10)$.
In the real-world production~\cite{derrow2021eta}, the forecasting models are usually updated in an online fashion~(\eg per hours).
Therefore, we estimate the missing latest traffic states based on previous input data, $\mathbfcal{\tilde{H}}_{t} = g_{\varphi}(\mathbfcal{H}_{t-1})$,
where $g_{\varphi}(\cdot)$ is the estimation function parameterized by $\varphi$. For simplicity, we directly obtain $\varphi$ from the pre-trained traffic forecasting model $f_{\theta^\ast}(\cdot)$.

With the surrogate traffic state label $\mathbf{\tilde{Y}}_{t+1:t+\tau}$, we derive the time-dependent node saliency~(TDNS) for each node as
\begin{equation}\label{eq:explanation_nodes}
    \mathcal{M}_{t} = \left \|  \sigma(\frac{\partial \mathcal{L}( f_{\theta}(\mathbfcal{\tilde{H}}_{t-T+1:t}),\mathbf{\tilde{Y}}_{t+1:t+\tau})}{\partial \mathbf{\tilde{X}}_{t-T+1:t}})\right \|_{p},
\end{equation}
where $\mathcal{L}( f_{\theta  }(\mathbfcal{\tilde{H}}_{t-T+1:t}),\mathbf{\tilde{Y}}_{t+1:t+\tau})$ is the  loss function and $\sigma$ is the activation function.
Intuitively, $\mathcal{M}_{t}$ reveals the node-wise loss impact with the same degree of perturbations.
Note depending on the time step $t$, $\mathcal{M}_{t}$ may vary.
A similar idea also has been adopted to identify static pixel saliency for image classification~\cite{simonyan2013deep}.

More in detail, the  loss function $\mathcal{L}( f_{\theta  }(\mathbfcal{\tilde{H}}_{t-T+1:t}),\mathbf{\tilde{Y}}_{t+1:t+\tau})$ in Equation~\ref{eq:explanation_nodes} is updated by the iterative gradient-based adversarial method~\cite{madry2018towards-PGD},
 \begin{equation}\label{eq:adversarial_loss_PGD}
    \mathbf{X'}_{t-T+1:t}^{(i)} =\text{clip}_{\mathbf{X'}_{t-T+1:t}, \varepsilon }(\mathbf{X'}_{t-T+1:t}^{(i-1)} + \alpha \text{sign}( \nabla \mathcal{L}(f_{\theta ^{\ast} }(\mathbfcal{H'}_{t-T+1:t}^{(i-1)}),\mathbf{\tilde{Y}}_{t+1:t+\tau}))),
\end{equation}
where $\mathbfcal{H'}_{t-T+1:t}^{(i)}$ is  adversarial traffic states at $i$-th iteration, $\alpha$ is the step size, and  $\text{clip}_{\mathbf{X'}_{t-T+1}, \varepsilon }(\cdot)$ is the project operation which clips the spatiotemporal feature with maximum perturbation bound $\varepsilon$. Note $\mathbfcal{H'}_{t-T+1:t}^{(0)}= \mathbfcal{\tilde{H}}_{t-T+1:t}$.

\eat{
For a batch of data $\left \{ (\mathbfcal{\tilde{H}}_{t-T+1:t}, \mathbf{\tilde{Y}}_{t+1:t+\tau})_{(j)} \right \}_{j=0}^{s-1}$,  we fuse their time-dependent
node saliency gradients,
\begin{equation}
        \mathbf{g}_{t} = \frac{1}{s}\sum_{j}\{\frac{\partial \mathcal{L}( f_{\theta ^{\ast} }(\mathbfcal{\tilde{H}}_{t-T+1:t} ), \mathbf{\tilde{Y}}_{t+1:t+\tau} )}{\partial \mathbf{X'}_{t-T+1:t}}\}_{j},
\end{equation}
where $s$ is the batch size.
To get the final time-dependent node saliency, we  use the RELU function to filter negative values  obtaining the positive feedback,
\begin{equation}\label{eq:node_saliency}
  \mathcal{M}_{t} = \left \| \text{Relu}(\mathbf{g}_{t}) \right \|_{2},
\end{equation}

The TNDS approach is described in algorithm \ref{alg:spatiotemporal-specific node saliency} in the Appendix.
After getting the time-dependent node saliency $\mathcal{M}_{t}$, we define a target node mask to compute the victim node selection indication $S_{t}$ in Eq. \ref{eq:target_node_mask},

\begin{equation}\label{eq:target_node_mask}
 s_{n,t}=\begin{cases}
 1 & \text{ if } n \in \text{Top}(\mathcal{M}_{t},\eta) \\
 0 & \text{ otherwise }, \end{cases}
\end{equation}
where $s_{n,t}$ is the $n$-th value of $S_{t}$ at time step $t$, and $\text{Top}(\cdot)$ return the  node set with  top $\eta$   spatiotemporal-specific node saliency.
}

For each batch of data $\left \{ (\mathbfcal{\tilde{H}}_{t-T+1:t}, \mathbf{\tilde{Y}}_{t+1:t+\tau})_{(j)} \right \}_{j=1}^{\gamma}$, the time-dependent node saliency gradient is derived by
\begin{equation}\label{eq:average_fuse_operation}
        \mathbf{g}_{t} = \frac{1}{\gamma}\sum_{j}\{\frac{\partial \mathcal{L}( f_{\theta ^{\ast} }(\mathbfcal{\tilde{H}}_{t-T+1:t} ), \mathbf{\tilde{Y}}_{t+1:t+\tau} )}{\partial \mathbf{X'}_{t-T+1:t}}\}_{j},
\end{equation}
where $\gamma$ is the batch size.
We use the RELU activation function to compute the non-negative saliency score for each time step,
\begin{equation}\label{eq:node_saliency}
  \mathcal{M}_{t} = \left \| \text{Relu}(\mathbf{g}_{t}) \right \|_{2}.
\end{equation}

Finally, we obtain the set of victim node $S_{t}$ based on $\mathcal{M}_{t}$,
\begin{equation}\label{eq:target_node_mask}
 s_{(i,i),t}=\begin{cases}
 1 & \text{ if } v_i \in \text{Top}(\mathcal{M}_{t},k) \\
 0 & \text{ otherwise }, \end{cases}
\end{equation}
where $s_{(i,i),t}$ denotes the $i$-th diagonal element of $S_t$, and $\text{Top}(\cdot)$ is a 0-1 indicator function returning if $v_i$ is the top-$k$ salient node at time step $t$.

\eat{
We will iterate  iterate $K$  steps to fully explore the sample space, then we fuse the final  time-dependent node saliency gradients with average operation,

\begin{equation}\label{eq:average_grad}
    \mathbf{g}_{t} = \frac{1}{s}\sum_{t}\frac{\partial \mathcal{L}( f_{\theta ^{\ast} }(\mathbfcal{H'}_{t-T+1:t} ,\tilde{Y}_{t+1:t+\tau})) }{\partial \mathbf{X'}_{t-T+1:t}};
\end{equation}
}

\subsection{Attack with adversarial traffic state}\label{sec:attack-traffic}
\eat{
 Then, we incorporate the interactive gradient-based attack  PGD with optimal victim node at each step to generate adversarial  spatiotemporal feature named spatiotemporal PGD (STPGD),
\begin{equation}\label{eq:white-box_ST-PGD}
    \mathbf{X'}_{t-T+1:t}^{(i)} =\text{clip}_{\mathbf{X'}_{t}, \varepsilon }(\mathbf{X'}_{t-T+1:t}^{(i-1)} + \alpha \text{sign}( \nabla \mathcal{L}(f_{\theta ^{\ast} }(\mathbfcal{H'}_{t-T+1:t}^{(i-1)}),\mathbf{\tilde{Y}}_{t+1:t+\tau})\cdot S_{t})),
\end{equation}
where $\mathbfcal{H'}_{t-T+1:t}^{(i-1)}= \left\{ (\mathbf{X'}_{t}^{(i-1)}, \mathbfcal{G}_{t})\right\}$ is the adversarial traffic state at $i$-th iteration, $\alpha$ is is the step size.  $\text{clip}_{\mathbf{X'}_{t}, \varepsilon }$ is the operation to bound the adversarial feature in a $\varepsilon$ ball, and $\mathbf{X'}_{t}^{(0)}= \mathbf{\tilde{X}}_{t}$. STPGD  only optimizes the victim node at each step to compute the optimal adversarial traffic feature instead of, like naive PGD~\cite{madry2018towards-PGD}, indiscriminately optimizing the total nodes. Similarly, we can incorporate other iterative gradient-based methods, such as MIM~\cite{dong2018boosting-MIM} \eg with indication $S_{t}$ named STMIM.
}

Based on the time-dependent victim set, we conduct adversarial attacks to spatiotemporal traffic forecasting models. Specifically, we first generate perturbed adversarial traffic features based on gradient descent methods.
Take the widely used Projected Gradient Descent~(PGD)~\cite{madry2018towards-PGD} for illustration, we construct Spatiotemporal Projected Gradient Descent~(STPGD) as below,
\begin{equation}\label{eq:white-box_ST-PGD}
    \mathbf{X'}_{t-T+1:t}^{(i)} =\text{clip}_{\mathbf{X'}_{t-T+1:t}, \varepsilon }(\mathbf{X'}_{t-T+1:t}^{(i-1)} + \alpha \text{sign}( \nabla \mathcal{L}(f_{\theta ^{\ast} }(\mathbfcal{H'}_{t-T+1:t}^{(i-1)}),\mathbf{\tilde{Y}}_{t+1:t+\tau})\cdot S_{t})),
\end{equation}
where $\mathbfcal{H'}_{t-T+1:t}^{(i-1)}$ is the adversarial traffic state at $i-1$-th iteration in the iterative gradient descent, $\alpha$ is the step size, and $\text{clip}_{\mathbf{X'}_{t-T+1:t}, \varepsilon }(\cdot)$ is the operation to bound adversarial features in a $\varepsilon$ ball. Note $\mathbf{X'}_{t}^{(0)}= \mathbf{\tilde{X}}_{t}$.
Instead of perturbing all nodes as in vanilla PGD, we only inject perturbations on selected victim nodes in $S_{t}$.
Similarly, we can generate perturbed adversarial traffic features by extending other gradient based methods, such as MIM~\cite{dong2018boosting-MIM}.

In the testing phase, we can inject the adversarial traffic states $\mathbfcal{H'}_{t-T+1:t} = \mathbfcal{H}_{t-T+1:t} + \bigtriangleup \mathbfcal{H'}_{t-T+1:t}$ to apply adversarial attack, where $ \bigtriangleup \mathbfcal{H}_{t}' + \mathbfcal{H}_{t}= \left \{ (\mathbf{X}_{t}' -\mathbf{X}_{t})\cdot S_{t}+ \mathbf{X}_{t},  \mathcal{G}_{t} \right \} \in \mathbfcal{H'}_{t-T+1:t}$ and $\bigtriangleup \mathbfcal{H'}_{t} = \left \{((\mathbf{X'}_{t} -\mathbf{X}_{t})\cdot S_{t}, 0):   \left \|  S_{t} \right \|_{0} \le \eta , \left \|   (\mathbf{X'}_{t} -\mathbf{X}_{t})\cdot S_{t} \right \|_{p} \le \varepsilon \right\} \in \bigtriangleup\mathbfcal{H'}_{t-T+1:t}$.
The details of the adversarial spatiotemporal attack framework under the grey-box setting is in algorithm~\ref{alg:semi-box spatiotemporal adversarial attack}.

The overall adversarial spatiotemporal attack can be easily extended to the white-box and black-box settings, which are detailed below.

\textbf{White-box attack}. Since the adversaries can fully access the data and labels under the white-box setting, we directly use the real ground truth traffic states to guide the generation of adversarial traffic states.
The detailed algorithm is introduced in Appendix~\ref{sec:Adversarial spatiotemporal attack under the white-box setting}.

\textbf{Black-box attack}. The most restrictive black-box setting assumes limited accessibility to the target model and labels. Therefore, we first employ a surrogate model, which can be learned from the training data or by querying the traffic forecasting service~\cite{yuan2021survey, dong2021query}. Then we generate adversarial traffic states based on the surrogate model to attack the targeted traffic forecasting model.
Please refer to Appendix~\ref{sec:Adversarial spatiotemporal_attack_under_the_black-box_setting} for more details.

\eat{
We theoretically analyze the adversarial spatiotemporal traffic forecasting attack framework.

\begin{myTheo}\label{the:the1}
Let $f_{\theta}(\mathcal{G}_{t})$ and $f_{\theta}(\mathcal{G'}_{t})$  's $L$-th embeddings are $\mathbf{h}^{(L)}$ and  $\mathbf{h'}^{(L)}$, respectively,  based on assumption of activation function $\sigma$   is   locally Lipschitz continuous, and the maximum degree in the graph is $C$, more details in the Appendix, then the upper bound loss for  adversarial traffic state satisfies,
\begin{equation*}
    \left \|\mathbf{h}^{(L)}- \mathbf{h'}^{(L)}\right \|^{2}_{2} \le  (K C \lambda)^{2L} \varepsilon^{2} \eta,
\end{equation*}
where  $\lambda$  and $K$ are constant.
\end{myTheo}

The theorem \ref{the:the1} provides a general upper bound loss which reveal the  attack mechanism. The physical meaning of theorem is that the performance of spatiotemporal traffic forecasting model would be influenced by the size of adversarial perturbations and the number of chosen target nodes, structure of spatiotemporal forecasting models (activation function),
The structure of data geographic information (degree) .
}

We conclude this section with the theoretical upper bound analysis of the proposed adversarial attack strategy.  In particular, we demonstrate the attack performance against the spatiotemporal traffic forecasting model is related to the number of chosen victim nodes, the budget of adversarial perturbations, as well as the traffic network topology.

\begin{myTheo}\label{the:the1}
Let $\mathbf{Z}^{(L)}=f_{\theta}(\mathbfcal{H}_{t-T+1:t})$ and $\mathbf{Z}^{\prime(L)}=f_{\theta}(\mathbfcal{H'}_{t-T+1:t})$ be the $L$-th layer embeddings of the forecasting model, the upper bound of the adversarial loss satisfies
\begin{equation*}
    \left \|\mathbf{Z}^{(L)}- \mathbf{Z'}^{(L)}\right \|^{2}_{2} \le  (\lambda \beta C )^{2L} \varepsilon^{2} \eta,
\end{equation*}
where $\lambda$ denotes maximum weight bound in all layers of the forecasting model, $\beta$ denotes parameter of the activation function in $f_{\theta}(\cdot)$, $C$ denotes the maximum degree of $\mathcal{G}$.
$\eta$ and $\varepsilon$ are the budget of number of victim nodes and perturbations, respectively.
\end{myTheo}
\emph{Proof}. Please refer to Appendix~\ref{sec:theorem}.

\begin{algorithm}\label{alg:semi-box spatiotemporal adversarial attack}
\caption{Adversarial spatiotemporal attack under the grey-box setting}
\KwIn{
\eat{data $\mathbfcal{H}_{t-2T+1:t-N}$, pre-trained spatiotemporal model $f_{\theta^{\ast}}(\cdot)$,  pre-trained traffic state prediction model $g_{\varphi}(\cdot)$, maximum bound $\varepsilon$, iterations $K$, budget $\eta$, batch size $s$.}
Previous traffic data, pre-trained spatiotemporal model $f_{\theta^{\ast}}(\cdot)$, pre-trained traffic state prediction model $g_{\varphi}(\cdot)$, maximum perturbation budget $\varepsilon$, victim node budget $\eta$, and iterations $K$.
}
\KwResult{
\eat{Adversarial traffic state perturbations $\bigtriangleup\mathbfcal{H'}_{t-T+1:t}$.}
Perturbed Adversarial traffic states $\mathbfcal{H'}_{t-T+1:t}$.}
\tcc{Step 1: Identify time-dependent victim nodes}
Estimate current traffic state $\mathbfcal{\tilde{H}}_{t-N+1:t}$ by function $g_{\varphi}(\cdot)$\;
Construct future traffic state's surrogate labels $\mathbf{\tilde{Y}}_{t+1:t+\tau}$ by Equation~\ref{eq:pseudo_label} \;
Compute the time-dependent node saliency $\mathcal{M}_{t}$ with $\mathbfcal{\tilde{H}}_{t-T+1:t}$ and $\mathbf{\tilde{Y}}_{t+1:t+\tau}$ by Equation~\ref{eq:explanation_nodes}-\ref{eq:node_saliency}\;
Obtain the victim node set $S_{t}$ by Equations~\ref{eq:target_node_mask}  \;
\tcc{Step 2: Attack with adversarial traffic state}
Initialize adversarial traffic state $\mathbfcal{H'}_{t-T+1:t}^{(0)} = \mathbfcal{\tilde{H}}_{t-T+1:t} $\;
{\For { $i=1$  to $K$ }
{
Generate perturbed adversarial features $\mathbf{{X}'}^{(i)}_{t-T+1:t}$ by Equation~\ref{eq:white-box_ST-PGD}\;
$\bigtriangleup \mathbfcal{H}^{\prime (i)}_{t-T+1:t} = ((\mathbf{X}^{\prime(i)}_{t-T+1:t} -\mathbf{\tilde{X} }_{t-T+1:t})\cdot S_{t}, 0)$\;
}}
Return $\mathbfcal{H'}_{t-T+1:t} = \mathbfcal{H}_{t-T+1:t} + \bigtriangleup \mathbfcal{H'}_{t-T+1:t}$.
\end{algorithm}

\section{Experiments}

\subsection{Experimental setup}
\textbf{Datasets}. We use two popular real-world datasets to demonstrate the effectiveness of the proposed adversarial attack framework.
(1) \emph{PEMS-BAY}~\cite{chen2001freeway} traffic dataset is derived from the California Transportation Agencies (CalTrans) Performance Measurement System (PeMS) ranging from January 1, 2017 to May 31, 2017. 325 traffic sensors in the Bay Area collect traffic data every 5 minutes.
(2) \emph{METR-LA}~\cite{DCRNN} is a traffic speed dataset collected from 207 Los Angeles County roadway sensors. The traffic speed is recorded every 5 minutes and ranges from March 1, 2012 to June 30, 2012.
For evaluation, all datasets are chronologically
ordered, we take the first 70\% for training, the following 10\% for validation, and the rest 20\% for testing.
The statistics of the two datasets are reported in Appendix~\ref{sec:data}.

\textbf{Baselines}.
In the current literature, few studies can be directly applied to the real-valued traffic forecasting attack setting.
To guarantee the fairness of comparison, we construct two-step baselines as below.
For victim node identification, we adopt random selection and use the topology-based methods~(\ie node degree and betweenness centrality~\cite{brandes2001faster-centrality}) to select victim nodes. We also employ PageRank (PR)~\cite{page1999pagerank} as the baseline to decide the set of victim nodes.
For adversarial traffic state generation, we adopt two widely used iterative gradient-based methods, PGD~\cite{madry2018towards-PGD} and MIM~\cite{dong2018boosting-MIM}, to generate adversarial perturbations.
In summary, we construct eight two-step baselines, PGD-Random, PGD-PR, PGD-Centrality, PGD-Degree, MIM-Random, MIM-PR, MIM-Centrality, and MIM-Degree. For instance, PGD-PR indicates first identifying victim nodes with PageRank and then applying adversarial noises with PGD.
Depending on the adversarial perturbation method, we compare two variants of our proposed framework, namely STPGD-TDNS and STMIM-TDNS.

\eat{
a straightforward way is to randomly choose or use the topology information of the graph to decide victim nodes, such as the node degree and betweenness centrality~\cite{brandes2001faster-centrality}. We also randomly select victim nodes and use PageRank (PR)~\cite{page1999pagerank} to rank nodes. The top-ranked nodes would be the victim nodes. we use iterative  gradient-based method including PGD~\cite{madry2018towards-PGD} and MIM~\cite{dong2018boosting-MIM}  to generate total nodes'  the adversarial noise, and finally add the corresponding noises to victim nodes.

The compared baselines include PGD-Random, PGD-PR, PGD-Centrality, PGD-Degree, MIM-Random, MIM-PR, MIM-Centrality, MIM-Degree. For example, PGD-Random means randomly picking nodes and using the PGD method to generate adversarial noise.  Our methods are named STPGD-TNDS and STMIM-TNDS.
}

\textbf{Target model}.
To evaluate the generalization ability of the proposed adversarial attack framework, we adopt the state-of-the-art spatiotemporal traffic forecasting model, GraphWaveNet~(Gwnet)~\cite{Graph_Wave_Net}, as the target model. Evaluation results on more target models are reported in Appendix~\ref{sec:other_experments}.

\textbf{Evaluation metrics}. Our evaluation focus on both the global and local effect of adversarial attacks on spatiotemporal models,
\begin{subequations}
\begin{align}
    &   \mathbb{E}_{t \in \mathcal{T}_{test}} \mathcal{L}(f_{\theta}(\mathbfcal{H'}_{t-T+1:t}),\mathbf{Y}_{t+1:t+\tau}) \label{eq:global},\\
    &    \mathbb{E}_{t \in \mathcal{T}_{test}} \mathcal{L}(f_{\theta}(\mathbfcal{H'}_{t-T+1:t}),f_{\theta}(\mathbfcal{H}_{t-T+1:t})) \label{eq:local},
\end{align}\normalsize
\end{subequations}
where $\mathcal{L}(\cdot)$ is a user-defined loss function.
Different from the majority target of adversarial attacks that are classification models~(\eg adversarial accuracy), traffic forecasting is defined as a regression task.
Therefore, we adopt Mean Average Error~(MAE)~\cite{willmott2005advantages} and Root Mean Square Error~(RMSE)~\cite{chai2014root} for evaluation.
More specifically, we define Global MAE~(G-MAE), Local MAE~(L-MAE), Global RMSE~(G-RMSE), Local RMSE~(L-RMSE) to evaluate the effect of adversarial attacks on traffic forecasting.
Please refer to Appendix~\ref{sec:eva_met} for detailed definitions of four metrics.

\textbf{Implementation details}.
All experiments are implemented with PyTorch and performed on a Linux server with 4 RTX 3090 GPUs.The traffic speed is  normalized to $[0,1]$.  The input length $T$ and output length $\tau$ are set to $12$.
We select   10\% nodes from the whole nodes as the victim nodes, and   $\varepsilon$ is set to $0.5$. The  batch size $\gamma$ is set to $64$. The iteration $K$ is set to $5$, and the step size $\alpha$ is set to $0.1$.

\subsection{Overall attack performance}
\eat{
\textbf{Grey-box attack.}
The experimental results can be seen in Table \ref{tab:semi-PeMS-MetrLA-GWnet}.
Generally, the grey-box attacks are more difficult than white-box attacks due to the inaccessibility of future information.  The adversarial traffic states generated by grey-box adversarial traffic forecasting attack still reduce the performance of forecasting models.  The results indicate that adversarial traffic states have a strong attack ability. When compared to a clean MAE, the GWnet, for example, reduces performance by around two times, and our methods outperform all baselines. For example, with our adversarial traffic state attack, GWnet's  GMAE is 6.1329, when the clean MAE is 1.9775.
}

Table~\ref{tab:semi-PeMS-MetrLA-GWnet} reports the overall attack performance of our proposed approach against the original forecasting model and eight baselines with respect to four metrics. Note larger value indicates better attack performance and worse forecasting accuracy.
Specifically, we can make the following observations.
First, the adversarial attack can significantly degrade the traffic forecasting performance. For example, our approach achieves $(67.79\%, 62.31\%)$ and $(19.88\%, 14.55\%)$ global performance degradation compared with the original forecasting results on \emph{PeMS-BAY} and \emph{METR-LA} dataset, respectively.
Second, our approach achieves the best attack performance against all baselines. In particular, STPGD-TDNS achieves $(15.80\%, 15.39\%)$ global performance improvement and $(23.35\%, 17.19\%)$ local performance improvement on the \emph{PeMS-BAY} dataset. Similarly, STMIM-TDNS achieves $(2.44\%, 2.00\%)$ global performance improvement and $(11.20\%, 2.70\%)$ local performance improvement on the METR-LA dataset.
Moreover, we observe STPGD-TDNS and STMIM-TDNS, two variants of our framework, respectively achieve the best attack performance on \emph{PeMS-BAY} and \emph{METR-LA} datasets, which further validate the superiority of our framework for flexibly integrate different adversarial perturbation methods.
Overall, our adversarial attack framework successfully disrupts the traffic forecasting model to make biased predictions.

\begin{table}[tbp]
\caption{Adversarial attack performance under the grey-box setting.}\label{tab:semi-PeMS-MetrLA-GWnet}
\begin{adjustbox}{width=\linewidth}
\begin{tabular}{c|cccc|cccc}
\hline
             & \multicolumn{4}{c|}{\emph{PeMS-BAY}}                                          & \multicolumn{4}{c}{\emph{METR-LA}}                                            \\ \hline
\diagbox{Attack methods}{Metrics}              & G-MAE           & L-MAE           & G-RMSE           & L-RMSE          & G-MAE           & L-MAE           & G-RMSE           & L-RMSE          \\ \hline
non-attack          & 1.975           & -               & 4.0220           & -               & 6.3504          & -               & 11.8424          & -               \\ \hline
PGD-Random          & 4.9876          & 3.7431          & 8.9343           & 7.8006          & 7.8947          & 2.7030          & 13.2749          & 5.9501          \\
PGD-PR              & 4.8599          & 3.5819          & 8.8215           & 7.6727          & 7.9003          & 2.7070          & 13.2669          & 5.9132          \\
PGD-Centrality      & 5.1640          & 3.9585          & 9.1369           & 8.0333          & 7.8554          & 2.7107          & 13.3100          & 5.9422          \\
PGD-Degree          & 4.9121          & 3.6675          & 8.8486           & 7.7263          & 7.9011          & 2.7316          & 13.3738          & 6.0661          \\ \hline
MIM-Random          & 5.3645          & 4.1739          & 9.7082           & 8.6825          & 7.7115          & 2.3793          & 13.1724          & 5.6882          \\
MIM-PR              & 5.2405          & 4.0286          & 9.5902           & 8.5600          & 7.7206          & 2.3774          & 13.1294          & 5.6548          \\
MIM-Centrality      & 5.5321          & 4.3820          & 9.9312           & 8.9331          & 7.7074          & 2.4255          & 13.2233          & 5.7498          \\
MIM-Degree          & 5.3500          & 4.1745          & 9.5808           & 8.5573          & 7.7026          & 2.3877          & 13.2570          & 5.8229          \\ \hline
\textbf{STPGD-TDNS} & \textbf{6.1329} & \textbf{5.1647} & \textbf{10.6723} & \textbf{9.7003} & 7.7191          & 2.6534          & 13.6693          & 6.6794          \\
\textbf{STMIM-TDNS} & 5.6706          & 4.7010          & 10.1336          & 9.1813          & \textbf{7.9381} & \textbf{2.8848} & \textbf{13.8592} & \textbf{6.9885} \\ \hline
\end{tabular}
\end{adjustbox}
\end{table}

\subsection{Ablation study }
\eat{We take the ablation study to evaluate the effectiveness of two steps in our framework. To evaluate step 1, we randomly select the victim node labeled \textit{w/o saliency}. To evaluate step 2, we use PGD to generate the total node's adversarial perturbations and then only add corresponding noise to victim nodes labeled \textit{w/o optimal noise}. The results are reported in Table \ref{tab:ablation_study}.}

Then we conduct ablation study on our adversarial attack framework. Due to page limit, we report the result of STPGD-TDNS on the \emph{PeMS-BAY} dataset.
We consider two variants of our approach:
(1) \emph{w/o TDNS} that randomly choose victim nodes to attack, and
(2) \emph{w/o STPGD} that apply vanilla PGD noise to selected victim nodes.
As reported in Table~\ref{tab:ablation_study}, we observe $(3.91\%, 6.28\%, 2.97\%, 3.60\%)$ and $(33.41\%, 52.45\%, 26.19\%, 32.97\%)$ attack performance degradation on four metrics by removing our proposed TDNS and STPGD module, respectively.
The above results demonstrate the effectiveness of the two-step framework. Moreover, we observe that the STPGD module plays a more important role in the adversarial spatiotemporal attack.

\begin{table}[tbp]
\centering
\caption{Ablation study on \emph{PeMS-BAY}. }\label{tab:ablation_study}
\begin{adjustbox}{width=0.6\linewidth}
\begin{tabular}{c|cccc}
\hline
                                 & G-MAE                         & L-MAE           & G-RMSE           & L-RMSE         \\ \hline
non-attack                       & 1.975                         & -               & 4.0220           & -               \\
w/o TDNS                     & 5.9024                        & 4.8595          & 10.364           & 9.3635          \\
w/o STPGD               & 4.5969                        & 3.3876          & 8.4572            & 7.2949          \\
STPGD-TDNS                       & \textbf{6.1329}               & \textbf{5.1647} & \textbf{10.6723} & \textbf{9.7003} \\ \hline
\end{tabular}
\end{adjustbox}
\end{table}

\subsection{Parameter sensitivity}
\begin{figure}[tbp]
\centering
\includegraphics[width=0.99\textwidth]{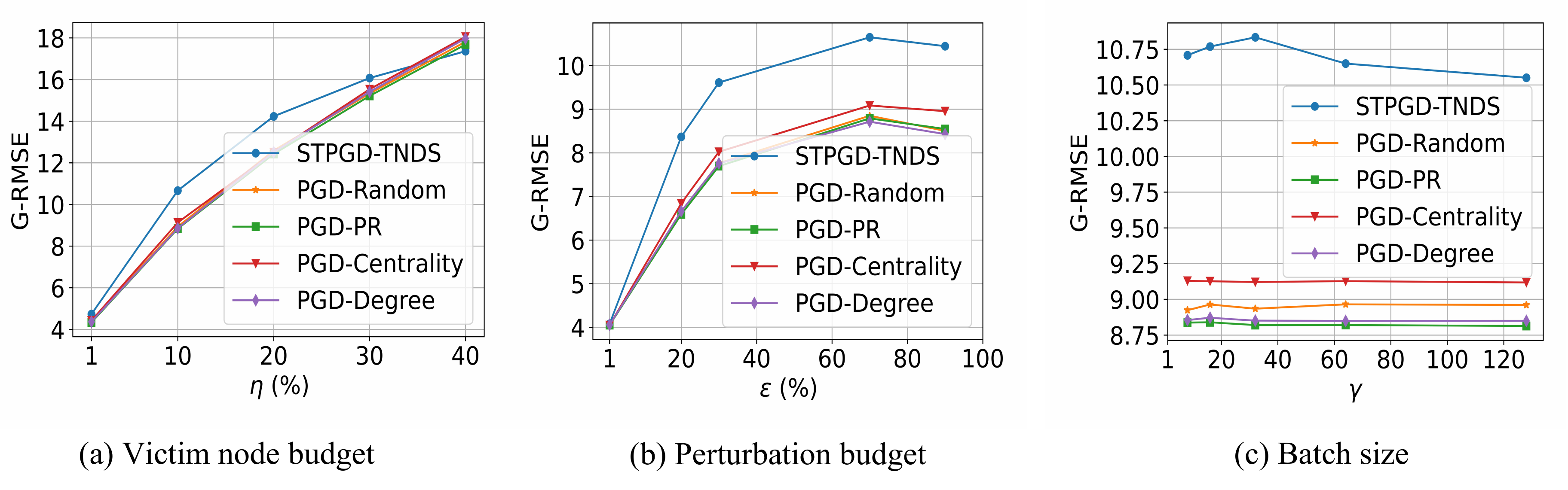}
\caption{Parameter sensitivity on \emph{PeMS-BAY}.}\label{fig:Impact_Epsilon_Eta}
\end{figure}

\eat{
We conduct experiments on \emph{PeMS-BAY} for Gwnet to analysis the core parameters sensitivity including budgets of perturbation $\varepsilon$ and victim node $\eta$, and batch size $s$ in our method.  The results are reported in Figure~\ref{fig:Impact_Epsilon_Eta}.

\textbf{Effect of perturbation bound}.
The number of victim nodes is fixed to $10\%$ total nodes and batch size $64$.
The G-RMSE first increase with the increase of  budget of perturbation $\varepsilon$, then decrease when $\varepsilon$ is 0.8.  In addition, our method always outperforms than other baselines with the increase of $\varepsilon$ in Figure~\ref{fig:Impact_Epsilon_Eta}~(a) on Gwnet.

\textbf{Effect of number of victim nodes budget}.
 We fix the maximum perturbation bound to $0.5$ and fusion batch size $64$, the G-RMSE rises as eta  increase from $1\%$ to $99\%$ total nodes in Figure~\ref{fig:Impact_Epsilon_Eta}~(b) on Gwnet. Although  the performance of our method is consistent with other baseline's after 40\% budget of victim nodes, our method maintains its superiority when the number of victim nodes is small.

\textbf{Effect of fusion batch size} We fix eta and epsilon to $10\%$ and $0.5$,  the G-RMSE decrease as fusion size  increase from $8$ to $128$  in Figure~\ref{fig:Impact_Epsilon_Eta}~(c)  We observe that there is a decrease with the batch size increase. Possibly, the large batch size brings more uncertainty when fusing the data.

}

We further study the parameter sensitivity of the proposed framework, including the number of victim nodes $\eta$, the perturbation budget $\varepsilon$, and the batch size $\gamma$.
Due to page limit, we report the result of G-RMSE on the \emph{PeMS-BAY} dataset. We observe similar results by using other metrics and on the \emph{METR-LA} dataset.
Each time we vary a parameter, we set other parameters to their default values.

\textbf{Effect of $\eta$}.
First, we vary the number of victim nodes from $0\%$ to $40\%$. As reported in Figure~\ref{fig:Impact_Epsilon_Eta}~(a), Our approach achieves the best attack performance with a limited victim node budget, and the advantage decrease when the attack can be applied to more nodes.

\textbf{Effect of $\varepsilon$}.
Second, we vary the perturbation budget from $0\%$ to $90\%$. As shown in Figure~\ref{fig:Impact_Epsilon_Eta}~(b), the G-RMSE first increase and then slightly decrease. This is perhaps because the clip function in Equation~\ref{eq:white-box_ST-PGD} weakens the diversity of attack noises.

\textbf{Effect of $\gamma$}.
Finally, we vary the batch size from $8$ to $128$, as illustrated in Figure~\ref{fig:Impact_Epsilon_Eta}~(c). We observe the adversarial attack is relatively stable to the batch size. However, too large batch size reduces the attack performance, which may induce over smooth of Equation~\ref{eq:average_fuse_operation}.

\subsection{Extended analysis under different attack settings}
Table~\ref{tab:white-black-PeMS-MetrLA-GWnet} reports the overall attack performance of our proposed approach against the original forecasting model and four PGD-based baselines under the white-box and black-box attack settings.
For the white-box attack, since the attacker can fully access the data and model, we re-train the forecasting model without requiring estimating the latest traffic states.
For the black-box attack, we adopt STAWNET~\cite{STAWnet} as the surrogate model.
The experimental results are summarized in Table~\ref{tab:white-black-PeMS-MetrLA-GWnet}.
First, we observe adversarial attacks significantly degrade the performance of the traffic forecasting model under both white-box and black-box settings.
For examples, our approach achieves $((68.65\%, 66.12\%)$ and $ (56.67\%, 50.78\%)$ global performance degradation compared with the vanilla forecasting model under white-box and black-box attack.
Moreover, our approach consistently achieves the best attack performance against baselines. To be more specific, our approach yield $(4.61\%,9.13\%)$ and $(1.70\%, 3.28\%)$ global performance improvement under the white-box setting and black-box setting, respectively.
In addition,  we observe higher attack effectiveness under the white-box setting and lower attack effectiveness under the black-box setting compared to the grey-box setting. This makes sense as the white-box setting can fully access the data and label, while the black-box has more restrictive data accessibility and relies on the surrogate model to apply adversarial spatiotemporal attack.

\begin{table}[tbp]
\caption{Adversarial attack performance on \emph{PeMS-BAY} under white-box and black-box settings.}\label{tab:white-black-PeMS-MetrLA-GWnet}
\begin{adjustbox}{width=\linewidth}
\begin{tabular}{c|cccc|cccc}
\hline
    & \multicolumn{4}{c|}{White-box}                                          & \multicolumn{4}{c}{Black-box}                                         \\ \hline
\diagbox{Attack methods}{Metrics}          & G-MAE           & L-MAE           & G-RMSE           & L-RMSE           & G-MAE           & L-MAE           & G-RMSE          & L-RMSE          \\ \hline
non-attack        & 2.0288          & -               & 4.2476           & -                & 1.9774          & -               & 4.0219          & -               \\ \hline
PGD-Random        & 6.1477          & 5.0463          & 10.9217          & 9.5163           & 4.241           & 2.9738          & 7.3804          & 5.99            \\
PGD-PR     & 6.1586          & 5.0713          & 10.7584          & 9.3405           & 4.4748          & 3.2605          & 7.9037          & 6.6306          \\
PGD-Centrality    & 6.1723          & 5.0823          & 10.9468          & 9.5272           & 4.4859          & 3.3002          & 7.8795          & 6.6045          \\
PGD-Degree        & 6.1507          & 5.0495          & 10.9375          & 9.5282           & 4.3577          & 3.1572          & 7.6159          & 6.2971          \\ \hline
\textbf{PGD-TDNS} & \textbf{6.4709} & \textbf{5.4953} & \textbf{12.1764} & \textbf{10.7262} & \textbf{4.5636} & \textbf{3.3543} & \textbf{8.1716} & \textbf{6.9388} \\ \hline
\end{tabular}
\end{adjustbox}
\end{table}

\subsection{Defense adversarial spatiotemporal attacks}
Finally, we study the defense of adversarial spatiotemporal attacks.
One primary goal of our study is to help improve the robustness of spatiotemporal forecasting models. Therefore, we propose to incorporate the adversarial training scheme for traffic forecasting models with our adversarial traffic states, denoted by \emph{AT-TNDS}. We compare it with (1) conventional adversarial training (\emph{AT})~\cite{madry2018towards-PGD} and (2) \emph{Mixup}~\cite{zhang2018mixup} with our adversarial traffic states. Note that we also tried other strategies, such as adding $L_{2}$ regularization, \etc, which fail to defend the adversarial spatiotemporal attack. The other state-of-the-art adversarial training methods, such as TRADE~\cite{tramer2019adversarial-TRADE}, cannot be directly applied in regression tasks.
Please refer to Appendix~\ref{sec:AT} for more training details.

The results in G-MAE on the \emph{PeMS-BAY} are reported in Table~\ref{tab:defense-gwnet}.
Overall, we observe \emph{AT} or \emph{Mixup} can successfully resist the adversarial spatiotemporal attack, and \emph{AT-TDNS} that combines the adversarial training scheme with our adversarial traffic states achieves the best defensive performance.
The above results indicate the defensibility of adversarial spatiotemporal attacks, which should be further investigated to deliver a more reliable spatiotemporal forecasting service in the future.

\eat{
\textbf{Defense adversarial traffic states}
We conduct further experiments to  evaluate the robustness of forecasting model with adversarial training~\footnote{Note that we also conduct other strategies, such as adding $L_{2}$ regularization methods \etal that fail to defend against adversarial attacks. The other SOTA adversarial training methods, such as TRADE~\cite{tramer2019adversarial-TRADE}, cannot be directly applied in regression tasks.}. We take the adversarial training including 1) adversarial training (\emph{AT})~\cite{madry2018towards-PGD}.  3) \emph{Mixup}~\cite{zhang2018mixup}.  3). We use adversarial data generated by our algorithm~\ref{alg:Adversarial_spatiotemporal_attack_under_the_white-box_setting} to train the model. The training details can be seen in Appendix.

The experimental results  are seen in Table \ref{tab:defense-gwnet}. The spatiotemporal traffic forecasting models trained with AT can defend adversarial traffic states compared to non defense strategy.  However, there still exits the performance degradation against spatiotemporal adversarial traffic attack. We  also find that adversarial training using our attacks may significantly improve the robustness of a spatiotemporal traffic forecasting models. For examples, in comparison to other defense strategies, we achieve SOTA adversarial robustness of the spatiotemporal traffic forecasting model.
}

\begin{table}[tbp]
\centering
\caption{Performance of defense adversarial spatiotemporal attack on \emph{PeMS-BAY}.  (
Values in parentheses indicate std)}\label{tab:defense-gwnet}
\begin{adjustbox}{width=0.99\linewidth}
\centering
\begin{tabular}{c|ccccc}
\hline
\diagbox{Defense strategies}{Attack methods}    & Non-attack      & PGD-Random      & PGD-PR & PGD-Centrality   & PGD-Degree      \\ \hline
Non-defense & \textbf{2.0288} & 6.1477          & 6.1586          & 6.1723          & 6.1507          \\ \hline
AT          & 2.1156          & 2.5436 (0.0249)          & 2.5539 (0.0375)         & 2.5660 (0.0281)          & 2.5394 (0.0279)          \\
Mixup       & 2.3090          & 2.7482 (0.0126)          & 2.7573 (0.0241)          & 2.7501 (0.0088)          & 2.7788 (0.0234)          \\
AT-TDNS     & 2.0935          & \textbf{2.4695 (0.0036)} & \textbf{2.4463 (0.0075) } & \textbf{2.4549 (0.0023)} & \textbf{2.4474 (0.0069)} \\ \hline
\end{tabular}
\end{adjustbox}
\end{table}

\section{Related work}

\textbf{Spatiotemporal traffic forecasting}. In recent years, the deep learning based traffic forecasting model has been extensively studied due to its superiority in jointly modeling temporal and spatial dependencies~\cite{STGCN, ASTGCN, AGCRN, T-GCN, Graph_Wave_Net, STAWnet, liu2021community, han2021joint}. To name a few,  STGCN~\cite{STGCN} applied graph convolution and gated causal convolution to capture the spatiotemporal information in the traffic domain, ASTGCN~\cite{ASTGCN} proposed a spatial-temporal attention network for capturing dynamic spatiotemporal correlations.
As another example, GraphWaveNet~\cite{Graph_Wave_Net} adaptively captures latent spatial dependency without requiring prior knowledge of the graph structure. The key objective of the above mentioned models is more accurate traffic forecasting. The vulnerability of spatiotemporal traffic forecasting models remains an under explored problem.

\textbf{Adversarial attack}. Deep neural networks have been proven vulnerable to adversarial examples~\cite{madry2018towards-PGD, goodfellow2014explaining}. As an emerging direction, various adversarial attack strategies on graph-structured data have been proposed, including both target-attack and non-target attack~\cite{jin2020adversarial, zhang2020adversarial}.
However, existing efforts on adversarial attacks mainly focus on classification tasks with static label~\cite{dong2018boosting-MIM, zhang2018mixup}.
Only a few works study the vulnerability of GCN based spatiotemporal forecasting models under query-based attack~\cite{zhu2021adversarial} and generate adversarial examples based on evolutionary algorithms~\cite{liu2021spatially}.
In this paper, we study the gradient based adversarial attack method against spatiotemporal traffic forecasting models, which is model-agnostic and generalizable to various attack settings, \ie white-box attack, grey-box attack, and black-box attack.

\section{Conclusion}

This paper showed the vulnerability of spatiotemporal traffic forecasting models under adversarial attacks. We proposed a practical adversarial spatiotemporal attack framework, which is agnostic to forecasting model architectures and is generalizable to various attack settings.
To be specific, we first constructed an iterative gradient guided node saliency method to identify a small time-dependent set of victim nodes.
Then, we proposed a spatiotemporal gradient descent based scheme to generate real-valued adversarial traffic states by flexibly leveraging various adversarial perturbation methods.
The theoretical analysis demonstrated the upper bound of the proposed two-step framework under human imperceptible victim node selection budget and perturbation budget constraints.
Finally, extensive experimental results on real-world datasets verify the effectiveness of the proposed framework.
The reported results will inspire further studies on the vulnerability of spatiotemporal forecasting models, as well as practical defending strategies for resisting adversarial attacks that can be deployed in real-world ITS systems.

\begin{ack}
This work is supported by the National Natural Science Foundation of China under Grant No.62102110, and Foshan HKUST Projects (FSUST21-FYTRI01A, FSUST21-FYTRI02A).
\end{ack}

{\small
\bibliographystyle{unsrt}

\begin{thebibliography}{10}

\bibitem{chen2021trafficstream}
Xu~Chen, Junshan Wang, and Kunqing Xie.
\newblock Trafficstream: {A} streaming traffic flow forecasting framework based
  on graph neural networks and continual learning.
\newblock In Zhi{-}Hua Zhou, editor, {\em Proceedings of the Thirtieth
  International Joint Conference on Artificial Intelligence, {IJCAI} 2021,
  Virtual Event / Montreal, Canada, 19-27 August 2021}, pages 3620--3626.
  ijcai.org, 2021.

\bibitem{Graph_Wave_Net}
Zonghan Wu, Shirui Pan, Guodong Long, Jing Jiang, and Chengqi Zhang.
\newblock Graph wavenet for deep spatial-temporal graph modeling.
\newblock In Sarit Kraus, editor, {\em Proceedings of the Twenty-Eighth
  International Joint Conference on Artificial Intelligence, {IJCAI} 2019,
  Macao, China, August 10-16, 2019}, pages 1907--1913. ijcai.org, 2019.

\bibitem{LiaoZWMCYGW18}
Binbing Liao, Jingqing Zhang, Chao Wu, Douglas McIlwraith, Tong Chen, Shengwen
  Yang, Yike Guo, and Fei Wu.
\newblock Deep sequence learning with auxiliary information for traffic
  prediction.
\newblock In Yike Guo and Faisal Farooq, editors, {\em Proceedings of the 24th
  {ACM} {SIGKDD} International Conference on Knowledge Discovery {\&} Data
  Mining, {KDD} 2018, London, UK, August 19-23, 2018}, pages 537--546. {ACM},
  2018.

\bibitem{qiu2019nei}
Jing Qiu, Lei Du, Dongwen Zhang, Shen Su, and Zhihong Tian.
\newblock Nei-tte: Intelligent traffic time estimation based on fine-grained
  time derivation of road segments for smart city.
\newblock {\em {IEEE} Trans. Ind. Informatics}, 16(4):2659--2666, 2020.

\bibitem{derrow2021eta}
Austin Derrow{-}Pinion, Jennifer She, David Wong, Oliver Lange, Todd Hester,
  Luis Perez, Marc Nunkesser, Seongjae Lee, Xueying Guo, Brett Wiltshire,
  Peter~W. Battaglia, Vishal Gupta, Ang Li, Zhongwen Xu, Alvaro
  Sanchez{-}Gonzalez, Yujia Li, and Petar Velickovic.
\newblock {ETA} prediction with graph neural networks in google maps.
\newblock In Gianluca Demartini, Guido Zuccon, J.~Shane Culpepper, Zi~Huang,
  and Hanghang Tong, editors, {\em {CIKM} '21: The 30th {ACM} International
  Conference on Information and Knowledge Management, Virtual Event,
  Queensland, Australia, November 1 - 5, 2021}, pages 3767--3776. {ACM}, 2021.

\bibitem{AGCRN}
Lei Bai, Lina Yao, Can Li, Xianzhi Wang, and Can Wang.
\newblock Adaptive graph convolutional recurrent network for traffic
  forecasting.
\newblock In Hugo Larochelle, Marc'Aurelio Ranzato, Raia Hadsell,
  Maria{-}Florina Balcan, and Hsuan{-}Tien Lin, editors, {\em Advances in
  Neural Information Processing Systems 33: Annual Conference on Neural
  Information Processing Systems 2020, NeurIPS 2020, December 6-12, 2020,
  virtual}, 2020.

\bibitem{xu2020adversarial}
Han Xu, Yao Ma, Haochen Liu, Debayan Deb, Hui Liu, Jiliang Tang, and Anil~K.
  Jain.
\newblock Adversarial attacks and defenses in images, graphs and text: {A}
  review.
\newblock {\em Int. J. Autom. Comput.}, 17(2):151--178, 2020.

\bibitem{madry2018towards-PGD}
Aleksander Madry, Aleksandar Makelov, Ludwig Schmidt, Dimitris Tsipras, and
  Adrian Vladu.
\newblock Towards deep learning models resistant to adversarial attacks.
\newblock In {\em 6th International Conference on Learning Representations,
  {ICLR} 2018, Vancouver, BC, Canada, April 30 - May 3, 2018, Conference Track
  Proceedings}. OpenReview.net, 2018.

\bibitem{dong2018boosting-MIM}
Yinpeng Dong, Fangzhou Liao, Tianyu Pang, Hang Su, Jun Zhu, Xiaolin Hu, and
  Jianguo Li.
\newblock Boosting adversarial attacks with momentum.
\newblock In {\em 2018 {IEEE} Conference on Computer Vision and Pattern
  Recognition, {CVPR} 2018, Salt Lake City, UT, USA, June 18-22, 2018}, pages
  9185--9193. Computer Vision Foundation / {IEEE} Computer Society, 2018.

\bibitem{STGCN}
Bing Yu, Haoteng Yin, and Zhanxing Zhu.
\newblock Spatio-temporal graph convolutional networks: {A} deep learning
  framework for traffic forecasting.
\newblock In J{\'{e}}r{\^{o}}me Lang, editor, {\em Proceedings of the
  Twenty-Seventh International Joint Conference on Artificial Intelligence,
  {IJCAI} 2018, July 13-19, 2018, Stockholm, Sweden}, pages 3634--3640.
  ijcai.org, 2018.

\bibitem{ASTGCN}
Shengnan Guo, Youfang Lin, Ning Feng, Chao Song, and Huaiyu Wan.
\newblock Attention based spatial-temporal graph convolutional networks for
  traffic flow forecasting.
\newblock In {\em The Thirty-Third {AAAI} Conference on Artificial
  Intelligence, {AAAI} 2019, The Thirty-First Innovative Applications of
  Artificial Intelligence Conference, {IAAI} 2019, The Ninth {AAAI} Symposium
  on Educational Advances in Artificial Intelligence, {EAAI} 2019, Honolulu,
  Hawaii, USA, January 27 - February 1, 2019}, pages 922--929. {AAAI} Press,
  2019.

\bibitem{STAWnet}
Chenyu Tian and Wai Kin~(Victor) Chan.
\newblock Spatial-temporal attention wavenet: A deep learning framework for
  traffic prediction considering spatial-temporal dependencies.
\newblock {\em IET Intelligent Transport Systems}, 15(4):549--561, 2021.

\bibitem{zhang2017deep}
Junbo Zhang, Yu~Zheng, and Dekang Qi.
\newblock Deep spatio-temporal residual networks for citywide crowd flows
  prediction.
\newblock In Satinder Singh and Shaul Markovitch, editors, {\em Proceedings of
  the Thirty-First {AAAI} Conference on Artificial Intelligence, February 4-9,
  2017, San Francisco, California, {USA}}, pages 1655--1661. {AAAI} Press,
  2017.

\bibitem{goodfellow2014explaining}
Ian~J. Goodfellow, Jonathon Shlens, and Christian Szegedy.
\newblock Explaining and harnessing adversarial examples.
\newblock In Yoshua Bengio and Yann LeCun, editors, {\em 3rd International
  Conference on Learning Representations, {ICLR} 2015, San Diego, CA, USA, May
  7-9, 2015, Conference Track Proceedings}, 2015.

\bibitem{simonyan2013deep}
Karen Simonyan, Andrea Vedaldi, and Andrew Zisserman.
\newblock Deep inside convolutional networks: Visualising image classification
  models and saliency maps.
\newblock In Yoshua Bengio and Yann LeCun, editors, {\em 2nd International
  Conference on Learning Representations, {ICLR} 2014, Banff, AB, Canada, April
  14-16, 2014, Workshop Track Proceedings}, 2014.

\bibitem{yuan2021survey}
Haitao Yuan and Guoliang Li.
\newblock A survey of traffic prediction: from spatio-temporal data to
  intelligent transportation.
\newblock {\em Data Science and Engineering}, 6(1):63--85, 2021.

\bibitem{dong2021query}
Yinpeng Dong, Shuyu Cheng, Tianyu Pang, Hang Su, and Jun Zhu.
\newblock Query-efficient black-box adversarial attacks guided by a
  transfer-based prior.
\newblock {\em IEEE Transactions on Pattern Analysis and Machine Intelligence},
  pages 1--1, 2021.

\bibitem{chen2001freeway}
Chao Chen, Karl Petty, Alexander Skabardonis, Pravin Varaiya, and Zhanfeng Jia.
\newblock Freeway performance measurement system: mining loop detector data.
\newblock {\em Transportation Research Record}, 1748(1):96--102, 2001.

\bibitem{DCRNN}
Yaguang Li, Rose Yu, Cyrus Shahabi, and Yan Liu.
\newblock Diffusion convolutional recurrent neural network: Data-driven traffic
  forecasting.
\newblock In {\em 6th International Conference on Learning Representations,
  {ICLR} 2018, Vancouver, BC, Canada, April 30 - May 3, 2018, Conference Track
  Proceedings}. OpenReview.net, 2018.

\bibitem{brandes2001faster-centrality}
Ulrik Brandes.
\newblock A faster algorithm for betweenness centrality.
\newblock {\em Journal of mathematical sociology}, 25(2):163--177, 2001.

\bibitem{page1999pagerank}
Lawrence Page, Sergey Brin, Rajeev Motwani, and Terry Winograd.
\newblock The pagerank citation ranking: Bringing order to the web.
\newblock Technical report, Stanford InfoLab, 1999.

\bibitem{willmott2005advantages}
Cort~J Willmott and Kenji Matsuura.
\newblock Advantages of the mean absolute error ({MAE}) over the root mean
  square error ({RMSE}) in assessing average model performance.
\newblock {\em Climate research}, 30(1):79--82, 2005.

\bibitem{chai2014root}
Tianfeng Chai and Roland~R Draxler.
\newblock Root mean square error ({RMSE}) or mean absolute error
  ({MAE})?--arguments against avoiding rmse in the literature.
\newblock {\em Geoscientific model development}, 7(3):1247--1250, 2014.

\bibitem{zhang2018mixup}
Hongyi Zhang, Moustapha Ciss{\'{e}}, Yann~N. Dauphin, and David Lopez{-}Paz.
\newblock mixup: Beyond empirical risk minimization.
\newblock In {\em 6th International Conference on Learning Representations,
  {ICLR} 2018, Vancouver, BC, Canada, April 30 - May 3, 2018, Conference Track
  Proceedings}. OpenReview.net, 2018.

\bibitem{tramer2019adversarial-TRADE}
Florian Tramer and Dan Boneh.
\newblock Adversarial training and robustness for multiple perturbations.
\newblock {\em Advances in Neural Information Processing Systems}, 32, 2019.

\bibitem{T-GCN}
Ling Zhao, Yujiao Song, Chao Zhang, Yu~Liu, Pu~Wang, Tao Lin, Min Deng, and
  Haifeng Li.
\newblock T-gcn: A temporal graph convolutional network for traffic prediction.
\newblock {\em IEEE Transactions on Intelligent Transportation Systems},
  21(9):3848--3858, 2020.

\bibitem{liu2021community}
Hao Liu, Qiyu Wu, Fuzhen Zhuang, Xinjiang Lu, Dejing Dou, and Hui Xiong.
\newblock Community-aware multi-task transportation demand prediction.
\newblock In {\em Proceedings of the AAAI Conference on Artificial
  Intelligence}, volume~35, pages 320--327, 2021.

\bibitem{han2021joint}
Jindong Han, Hao Liu, Hengshu Zhu, Hui Xiong, and Dejing Dou.
\newblock Joint air quality and weather prediction based on multi-adversarial
  spatiotemporal networks.
\newblock In {\em Proceedings of the AAAI Conference on Artificial
  Intelligence}, volume~35, pages 4081--4089, 2021.

\bibitem{jin2020adversarial}
Wei Jin, Yaxin Li, Han Xu, Yiqi Wang, Shuiwang Ji, Charu Aggarwal, and Jiliang
  Tang.
\newblock Adversarial attacks and defenses on graphs: A review, a tool and
  empirical studies.
\newblock {\em arXiv preprint arXiv:2003.00653}, 2020.

\bibitem{zhang2020adversarial}
Zijie Zhang, Zeru Zhang, Yang Zhou, Yelong Shen, Ruoming Jin, and Dejing Dou.
\newblock Adversarial attacks on deep graph matching.
\newblock {\em Advances in Neural Information Processing Systems},
  33:20834--20851, 2020.

\bibitem{zhu2021adversarial}
Lyuyi Zhu, Kairui Feng, Ziyuan Pu, and Wei Ma.
\newblock Adversarial diffusion attacks on graph-based traffic prediction
  models.
\newblock {\em arXiv preprint arXiv:2104.09369}, 2021.

\bibitem{liu2021spatially}
Fuqiang Liu, Luis Miranda-Moreno, and Lijun Sun.
\newblock Spatially focused attack against spatiotemporal graph neural
  networks.
\newblock {\em arXiv preprint arXiv:2109.04608}, 2021.

\bibitem{MTGNN}
Zonghan Wu, Shirui Pan, Guodong Long, Jing Jiang, Xiaojun Chang, and Chengqi
  Zhang.
\newblock Connecting the dots: Multivariate time series forecasting with graph
  neural networks.
\newblock In Rajesh Gupta, Yan Liu, Jiliang Tang, and B.~Aditya Prakash,
  editors, {\em {KDD} '20: The 26th {ACM} {SIGKDD} Conference on Knowledge
  Discovery and Data Mining, Virtual Event, CA, USA, August 23-27, 2020}, pages
  753--763. {ACM}, 2020.

\end{thebibliography}

}

\section*{Checklist}

\begin{enumerate}

\item For all authors...
\begin{enumerate}
  \item Do the main claims made in the abstract and introduction accurately reflect the paper's contributions and scope?
  \answerYes{}
  \item Did you describe the limitations of your work?
\answerYes{}
  \item Did you discuss any potential negative societal impacts of your work?
   \answerNA{}
  \item Have you read the ethics review guidelines and ensured that your paper conforms to them?
  \answerYes{}
\end{enumerate}

\item If you are including theoretical results...
\begin{enumerate}
  \item Did you state the full set of assumptions of all theoretical results?
    \answerYes{}
	\item Did you include complete proofs of all theoretical results?
        \answerYes{}
\end{enumerate}

\item If you ran experiments...
\begin{enumerate}
  \item Did you include the code, data, and instructions needed to reproduce the main experimental results (either in the supplemental material or as a URL)?
        \answerYes{ }
  \item Did you specify all the training details (e.g., data splits, hyperparameters, how they were chosen)?
  \answerYes{}
	\item Did you report error bars (e.g., with respect to the random seed after running experiments multiple times)?
    \answerYes{}
	\item Did you include the total amount of compute and the type of resources used (e.g., type of GPUs, internal cluster, or cloud provider)?
    \answerYes{ }
\end{enumerate}

\item If you are using existing assets (e.g., code, data, models) or curating/releasing new assets...
\begin{enumerate}
  \item If your work uses existing assets, did you cite the creators?
      \answerYes{ }
  \item Did you mention the license of the assets?
 \answerYes{ }
  \item Did you include any new assets either in the supplemental material or as a URL?
   \answerYes{}
  \item Did you discuss whether and how consent was obtained from people whose data you're using/curating?
     \answerYes{}
  \item Did you discuss whether the data you are using/curating contains personally identifiable information or offensive content?
      \answerYes{}
\end{enumerate}

\item If you used crowdsourcing or conducted research with human subjects...
\begin{enumerate}
  \item Did you include the full text of instructions given to participants and screenshots, if applicable?
     \answerNA{}
  \item Did you describe any potential participant risks, with links to Institutional Review Board (IRB) approvals, if applicable?
       \answerNA{}
  \item Did you include the estimated hourly wage paid to participants and the total amount spent on participant compensation?
     \answerNA{}
\end{enumerate}

\end{enumerate}


\clearpage
\appendix

\section{Adversarial spatiotemporal attack under different settings}

\subsection{Adversarial spatiotemporal attack under the white-box setting}\label{sec:Adversarial spatiotemporal attack under the white-box setting}
Since the adversaries can fully access the data and label under the white-box
setting, we directly use the real ground truth traffic states to generate the adversarial traffic states, as detailed in algorithm \ref{alg:Adversarial_spatiotemporal_attack_under_the_white-box_setting}.

\begin{algorithm}\label{alg:Adversarial_spatiotemporal_attack_under_the_white-box_setting}
\caption{Adversarial spatiotemporal attack under the white-box setting}
\KwIn{
\eat{data $\mathbfcal{H}_{t-2T+1:t-N}$, pre-trained spatiotemporal model $f_{\theta^{\ast}}(\cdot)$,  pre-trained traffic state prediction model $g_{\varphi}(\cdot)$, maximum bound $\varepsilon$, iterations $K$, budget $\eta$, batch size $s$.}
Previous traffic data, pre-trained spatiotemporal model $f_{\theta^{\ast}}(\cdot)$, pre-trained traffic state prediction model $g_{\varphi}(\cdot)$, maximum perturbation budget $\varepsilon$, victim node budget $\eta$, and iterations $K$.
}
\KwResult{
\eat{Adversarial traffic state perturbations $\bigtriangleup\mathbfcal{H'}_{t-T+1:t}$.}
Perturbed adversarial traffic states $\mathbfcal{H'}_{t-T+1:t}$.}
\tcc{Step 1: Identify time-dependent victim nodes}
Compute the time-dependent node saliency $\mathcal{M}_{t}$ with $\mathbfcal{H}_{t-T+1:t}$ and $\mathbf{Y}_{t+1:t+\tau}$ by Equations~\ref{eq:explanation_nodes}-\ref{eq:node_saliency}\;
Obtain the victim node set $S_{t}$ by Equation~\ref{eq:target_node_mask}  \;
\tcc{Step 2: Attack with adversarial traffic state}
Initialize adversarial traffic state $\mathbfcal{H'}_{t-T+1:t}^{(0)} = \mathbfcal{H}_{t-T+1:t} $\;
{\For { $i=1$  to $K$ }
{
Generate perturbed adversarial features $\mathbf{{X}'}^{(i)}_{t-T+1:t}$ by Equation~\ref{eq:white-box_ST-PGD}\;

$\bigtriangleup \mathbfcal{H}^{\prime (i)}_{t-T+1:t} = ((\mathbf{X}^{\prime (i)}_{t-T+1:t} -\mathbf{X}^{(i)}_{t-T+1:t})\cdot S_{t}, 0)$\;

}}
Return $\mathbfcal{H'}_{t-T+1:t} = \mathbfcal{H}_{t-T+1:t} + \bigtriangleup \mathbfcal{H'}_{t-T+1:t}$.
\end{algorithm}

\subsection{Adversarial spatiotemporal attack under the black-box setting}\label{sec:Adversarial spatiotemporal_attack_under_the_black-box_setting}
The most restrictive black-box setting assumes limited accessibility to the target model and labels. Therefore, we first employ a surrogate model, which can be learned on the training data or query the traffic forecasting service~\cite{yuan2021survey, dong2021query}. Then we generate adversarial traffic states based on the surrogate model to attack the target model. In details, we use a surrogate model to generate the adversarial traffic states based on algorithm~\ref{alg:semi-box spatiotemporal adversarial attack}, the generated adversarial traffic states can be used to attack the target model.

\section{Proof}\label{sec:theorem}
In this section, we show the details of the proof. First, we recall the assumptions as follows:
let the $k$-th layer embedding of spatiotemporal traffic models is
\begin{subequations}
\begin{align}
    &\mathbf{Z}_{i}^{(k+1)}  = \sigma (\sum_{j\in \mathcal{N}_{i}}e_{ij}\mathbf{M}_{j}^{(k)}),
\end{align}\normalsize
\end{subequations}
where $\mathbf{Z}_{i}^{(k+1)}$ ($\mathbf{Z^{(0)}}=\mathcal{H}_{t-T+1:t}$) represents the embedding of node $v_i$ in $k+1$-th layer of the spatiotemporal forecasting model. $\mathbf{M}_{i}^{(k)} = \mathbf{Z}_{i}^{(k)}\mathbf{W}^{(k)}$, where $\mathbf{W}^{(k)}$ denotes the weight matrix  for $k$-th layer of the forecasting model.  $\sigma$ is an activation function, such as the sigmoid function, relu function, \etc $e_{ij}$ is the weight value used to aggregate node $j$'s neighbors. $\mathcal{N}_{i}$ represents the index used to keep track of node $j$'s neighbors. Let $\lambda$ denotes maximum weight bound in all layers of the forecasting model, where $\max_{k} \left \| \mathbf{W}^{(k)} \right \|_{2} \le \lambda, \forall k \in \left \{ 1, \cdots,L \right \}  $. We denote that the maximum degree in graph $\mathcal{G}$ is $C$.

\begin{myAss}\label{ass:ass1}
The activation function $\sigma$  used in spatiotemporal traffic forecasting model   is   locally Lipschitz continuous as,
\begin{equation}
 \left \|\sigma (\sum_{j\in \mathcal{N}_{i}}e_{ij}\mathbf{M}_{j}^{(k)})- \sigma (\sum_{j\in \mathcal{N}_{i}}e_{ij}\mathbf{M'}_{j}^{(k)})\right \|_{2} \le \beta \left \| \sum_{j\in \mathcal{N}_{i}}e_{ij}\mathbf{M}_{j}^{(k)} - \sum_{j\in \mathcal{N}_{i}}e_{ij}\mathbf{M'}_{j}^{(k)}\right \|_{p},
\end{equation}
where $\beta$ denotes parameter of the activation function in $f_{\theta}(\cdot)$.
\end{myAss}

\begin{proof}
\begin{subequations}
\begin{align*}
    &\left \|\mathbf{Z}^{(L)}- \mathbf{Z'}^{(L)}\right \|^{2}_{2} = \sum_{i}\left \|\mathbf{Z}^{(L)}_{i}- \mathbf{Z'}^{(L)}_{i}\right \|^{2}_{2}\\
    &\le \beta^{2}\sum_{i}\left \| \sum_{j\in \mathcal{N}_{i}}e_{ij} \mathbf{M}_{j}^{(L-1)}  - \sum_{j\in \mathcal{N}_{i}}e_{ij} \mathbf{M'}_{j}^{(L-1)}\right \|^{2}_{2}  \\
    &\le \beta^{2}\sum_{i} |\mathcal{N}_i| \sum_{j\in \mathcal{N}_{i}}\left \| e_{ij}(\mathbf{M}_{j}^{(L-1)}  - \mathbf{M'}_{j}^{(L-1)})\right \|^{2}_{2}  \\
    &\le \beta^2 C \sum_{i} \sum_{j\in \mathcal{N}_{i}}\left \|\mathbf{M}_{j}^{(L-1)}  - \mathbf{M'}_{j}^{(L-1)}\right \|^{2}_{2} \\
    & = \beta^{2} C \sum_{j}\left | \mathcal{N}_{j} \right |  \left \| \mathbf{M}_{j}^{(L-1)}  - \mathbf{M'}_{j}^{(L-1)}\right \|^{2}_{2}\\
    & \le \beta^{2} C^2 \sum_{j} \left \| \mathbf{M}_{j}^{(L-1)}  - \mathbf{M'}_{j}^{(L-1)}\right \|^{2}_{2}\\
    & \le (\beta C \lambda)^{2} \sum_{j} \left \| \mathbf{Z}_{j}^{(L-1)}  - \mathbf{Z'}_{j}^{(L-1)}\right \|^{2}_{2}\\
    & \le (\beta C \lambda)^{2L} \sum_{i} \left \| \mathbf{Z}_{i}^{(0)}  - \mathbf{Z'}_{i}^{(0)}\right \|^{2}_{2}\\
    &= (\beta C \lambda)^{2L}  (\sum_{ \text{i is the victim node}}\left \|  \mathbfcal{H}_{(i),(t-T+1:t)}  -  \mathbfcal{H}_{(i),(t-T+1:t)}'\right \|_{2}^{2}   \\
    &+ \sum_{ \text{i is not the victim node}}\left \|  \mathbfcal{H}_{(i),(t-T+1:t)}  -  \mathbfcal{H}_{(i),(t-T+1:t)}'\right \|_{2}^{2}  )\\
    & \le (\beta C \lambda)^{2L} \varepsilon^{2} \eta
\end{align*}\normalsize
\end{subequations}\label{eq:ST_proof1}
\end{proof}

\textbf{Remarks}. Assumption \ref{ass:ass1} provides a more general activation function assumption. This assumption is met by the ReLU, sigmoid, tanh function~\cite{STGCN, ASTGCN, Graph_Wave_Net} \etc  We also noticed that~\cite{zhu2021adversarial} also analyzes traffic forecasting loss under query-based attack. Our theorem is different in that we first give the worst performance bound of an adversarial traffic forecasting attack, but \cite{zhu2021adversarial} does not provide the worst performance bound. Second, our theorem is more general because we do not specify a specific activation function.

\section{Data statistics}\label{sec:data}
We conclude the data statistics for two-real world datasets in Table~\ref{tab:data}.

\begin{table}[htbp]
\caption{Data statistics}\label{tab:data}
\centering
\begin{tabular}{cccc}
\hline
Data     & Sample & Nodes & Traffic events \\ \hline
\emph{PeMS-BAY} & 34,272  & 325   & 16,937,700       \\
\emph{METR-LA}  & 52,116  & 207   & 7,094,304        \\ \hline
\end{tabular}
\end{table}

\section{Evaluation metric}\label{sec:eva_met}
 The Global MAE~(G-MAE), Local MAE~(L-MAE), Global RMSE~(G-RMSE), Local RMSE~(L-RMSE) are defined in Equations~\ref{eq:global_mae}-\ref{eq:local_rmse}.
\begin{subequations}
\begin{align}
    &       \text{G-MAE} = \frac{1}{m\times n}\sum_{t}\left \| f_{\theta}(\mathbfcal{H'}_{t-T+1:t})- \mathbf{Y}_{t+1:t+\tau} \right \|   \label{eq:global_mae}\\
    &       \text{L-MAE} = \frac{1}{m\times n}\sum_{t}\left \| f_{\theta}(\mathbfcal{H'}_{t-T+1:t}) -f_{\theta}(\mathbfcal{H}_{t-T+1:t})\right \|   \label{eq:local_mae},
\end{align}\normalsize
\end{subequations}

\begin{subequations}
\begin{align}
    &  \text{G-RMSE} =   \sqrt{\frac{1}{m\times n}\sum_{t}\left \| f_{\theta}(\mathbfcal{H'}_{t-T+1:t})- \mathbf{Y}_{t+1:t+\tau} \right \| ^{2} }\label{eq:global_rmse} \\
    &     \text{L-RMSE} = \sqrt{\frac{1}{m\times n}\sum_{t}\left \| f_{\theta}(\mathbfcal{H'}_{t-T+1:t}) -f_{\theta}(\mathbfcal{H}_{t-T+1:t}) \right \| ^{2} }  \label{eq:local_rmse},
\end{align}\normalsize
\end{subequations}
where $m$ represents the number of samples in test sets, and $n$ denotes the number of nodes.

\section{Defense adversarial traffic states}\label{sec:AT}
Given a spatiotemporal forecasting model $f_{\theta}(\cdot)$, the adversarial training in  spatiotemporal traffic forecasting is defined as
\begin{equation}
    \min_{\theta }\max_{\substack{\mathbfcal{H}^{\prime}_{t-T+1:t} \\ t \in \mathcal{T}_{train}}}
    \sum_{t \in \mathcal{T}_{train}} \mathcal{L}(f_{\theta }(\mathbfcal{H}^{\prime}_{t-T+1:t} ), \mathbf{Y}_{t+1:t+\tau}),
\end{equation}
where  $\mathcal{L}(\cdot)$ is the loss function measuring the distance between the predicted traffic states and ground truth, and $\theta$ is  parameters learned during the training stage.
$\mathcal{T}_{train}$ denote the set of time steps of all training samples.
We use strategies that include
(1) adversarial training (AT)~\cite{madry2018towards-PGD}. We use adversarial training with the PGD-Random adversarial attack method to generate the adversarial samples under white-box setting.
(2) Mixup~\cite{zhang2018mixup}. We randomly sample the clean  and adversarial samples to train the forecasting model. The  adversarial sample are also  generated by PGD-Random method under white-box setting.
(3). We use adversarial sampels generated by our method STPGD-TDNS under white-box setting to train the model.

\section{Further experiments}\label{sec:other_experments}
\subsection{Experiments on other models}

The other spatiotemporal traffic forecasting models are summarized as follows.
 (1) STGCN~\cite{STGCN} applies graph convolution and gated causal convolution to capture the spatiotemporal information in the traffic domain.
(2) To overcome the spatiotemporal forecasting problem, ASTGCN~\cite{ASTGCN} presented a spatial-temporal attention method for capturing dynamic spatiotemporal correlations. (3) MTGNN~\cite{MTGNN}  created a self-learned node embedding for forecasting traffic conditions that is also not dependent on a pre-defined graph.

We report the  evaluation results on other target models in Tables~\ref{tab:semi-PeMS-stgcn}-\ref{tab:semi-PeMS-mtgnn}. By carefully selecting victim nodes, the attacker can achieve more effective attack performance with less attack budget.   In particular, STPGD-TDNS achieves (62.23.80\%, 55.86\%) global performance improvement and (66.95.35\%, 59.25\%)
local performance improvement on the PeMS-BAY dataset for MTGNN.

\begin{table}[htbp]
\caption{Grey-box attack on STGCN for \emph{PeMS-BAY} }\label{tab:semi-PeMS-stgcn}
\centering
\begin{adjustbox}{width=0.6\linewidth}
\begin{tabular}{c|cccc}
\hline
\diagbox{Methods}{}              & G-MAE            & L-MAE           & G-RMSE           & L-RMSE           \\ \hline
non-attack          & 2.8324           & -               & 5.1708           & -                \\ \hline
PGD-Random          & 5.7924           & 4.0880          & 9.5659           & 8.0560           \\
PGD-PR              & 9.6118           & 8.1697          & 15.4945          & 14.6314          \\
PGD-Centrality      & 6.9712           & 5.1407          & 11.9507          & 10.7645          \\
PGD-Degree          & 6.3903  & 4.3974& 11.8196& 10.6630 \\ \hline
MIM-Random          & 6.0461           & 4.4043          & 9.8926           & 8.4604           \\
MIM-PR              & 9.5573           & 8.1512          & 15.2504          & 14.3865          \\
MIM-Centrality      & 6.9748           & 5.1906          & 11.6700          & 10.4777          \\
MIM-Degree          & 6.5071           & 4.5425          & 11.8073          & 10.6640          \\ \hline
\textbf{STPGD-TDNS} & 9.3440           & 7.8039          & 5.1708           & 14.8150          \\
\textbf{STMIM-TDNS} & \textbf{10.2563} & \textbf{8.7318} & \textbf{5.1708}  & \textbf{15.0358} \\ \hline
\end{tabular}
\end{adjustbox}
\end{table}

\begin{table}[htbp]
\caption{Grey-box attack on ASTGCN for \emph{PeMS-BAY} }\label{tab:semi-PeMS-MetrLA-astgcn}
\centering
\begin{adjustbox}{width=0.6\linewidth}
\begin{tabular}{c|cccc}
\hline
\diagbox{Methods}{}              & \multicolumn{1}{c}{G-MAE} & \multicolumn{1}{c}{L-MAE} & \multicolumn{1}{c}{G-RMSE} & \multicolumn{1}{c}{L-RMSE} \\ \hline
non-attack          & 2.3581                    & \multicolumn{1}{c}{-}     & 4.9165                     & \multicolumn{1}{c}{-}     \\ \hline
PGD-Random          & 5.2302                    & 3.1082                    & 11.5757                    & 10.4736                   \\
PGD-PR              & 5.2565                    & 3.1282                    & 11.6177                    & 10.5154                   \\
PGD-Centrality      & 5.2260                    & 3.1101                    & 11.5842                    & 10.4797                   \\
PGD-Degree          & 5.2504                    & 3.1377                    & 11.6332                    & 10.5305                   \\ \hline
MIM-Random          & 5.1907                    & 3.0609                    & 11.4680                    & 10.3509                   \\
MIM-PR              & 5.2080                    & 3.0787                    & 11.5024                    & 10.3861                   \\
MIM-Centrality      & 5.1733                    & 3.0569                    & 11.4584                    & 10.3409                   \\
MIM-Degree          & 5.2042                    & 3.0900                    & 11.5236                    & 10.4065                   \\ \hline
\textbf{STPGD-TDNS} & 5.2635                  & 3.1476                    & 11.6880                    & 10.5896                   \\
\textbf{STMIM-TDNS} & \textbf{5.2929}           & \textbf{3.1799}           & \textbf{11.7534}           & \textbf{10.6579}          \\ \hline
\end{tabular}
\end{adjustbox}
\end{table}

\begin{table}[htbp]
\caption{Grey-box attack on MTGNN for \emph{PeMS-BAY} }\label{tab:semi-PeMS-mtgnn}
\centering
\begin{adjustbox}{width=0.6\linewidth}
\begin{tabular}{c|cccc}
\hline
\diagbox{Methods}{}             & G-MAE            & L-MAE            & G-RMSE           & L-RMSE           \\ \hline
non-attack          & 2.1501           & -                & 4.2637           & -                \\ \hline
PGD-Random          & 5.4748           & 4.6839           & 9.5824           & 8.7328           \\
PGD-PR              & 4.7997           & 3.8990           & 8.7011           & 7.7349           \\
PGD-Centrality      & 5.6504           & 4.8921           & 9.6820           & 8.8529           \\
PGD-Degree          & 4.9282           & 4.0396           & 8.8791           & 7.9403           \\ \hline
MIM-Random          & 5.7671           & 4.9483           & 9.9446           & 9.1007           \\
MIM-PR              & 4.8927           & 3.9385           & 8.9900           & 8.0265           \\
MIM-Centrality      & 5.6832           & 4.8927           & 9.8080           & 8.9533           \\
MIM-Degree          & 4.9599           & 4.0260           & 9.0387           & 8.0839           \\ \hline
\textbf{STPGD-TDNS} & 14.9606          & 14.8017          & 21.9354          & 21.7272          \\
\textbf{STMIM-TDNS} & \textbf{16.0254} & \textbf{15.9020} & \textbf{23.3589} & \textbf{23.1604} \\ \hline
\end{tabular}
\end{adjustbox}
\end{table}

\subsection{Ablation study  under white-box setting}
Since selecting a few set as the victim nodes is   important to attack traffic forecasting model, we conduct further ablation study to evaluate the method TDNS under the white-box setting. Table \ref{tab:white-PeMS-MetrLA-GWnet-ST-AB} reports the overall results on Gwnet under white-box attack.
\begin{table}[htbp]
\centering
\caption{Ablation study under white-box attack on Gwnet for \emph{PeMS-BAY} }\label{tab:white-PeMS-MetrLA-GWnet-ST-AB}
\begin{adjustbox}{width=0.6\linewidth}
\begin{tabular}{c|cccc}
\hline
\diagbox{Methods}{}             & G-MAE           & L-MAE           & G-RMSE           & L-RMSE            \\ \hline
non-attack          & 2.0288          & -               & 4.2476           & -                \\ \hline
STPGD-Random        & 6.1477          & 5.0463          & 10.9217          & 9.5163           \\
STPGD-PR            & 6.1586          & 5.0713          & 10.7584          & 9.3405           \\
STPGD-Centrality    & 6.1723          & 5.0823          & 10.9468          & 9.5272           \\
STPGD-Degree        & 6.1507          & 5.0495          & 10.9375          & 9.5282           \\ \hline
STMIM-Random        & 5.9524          & 4.8091          & 10.6488          & 9.1917           \\
STMIM-PR            & 5.9311          & 4.7954          & 10.4354          & 8.9565           \\
STMIM-Centrality    & 5.9159          & 4.7786          & 10.5948          & 9.1180           \\
STMIM-Degree        & 5.9570          & 4.8085          & 10.6692          & 9.2136           \\ \hline
\textbf{STPGD-TDNS} & \textbf{6.4709} & \textbf{5.4953} & \textbf{12.1764} & \textbf{10.7262} \\
\textbf{STMIM-TDNS} & 6.3018          & 5.2733          & 11.8618          & 10.3729          \\ \hline
\end{tabular}
\end{adjustbox}
\end{table}

\eat{
\begin{table}[htbp]
\centering
\caption{Ablation study under grey-box on Gwnet for PeMS}\label{tab:semi-PeMS-MetrLA-GWnet-ST-AB}
\begin{adjustbox}{width=0.6\linewidth}
\begin{tabular}{c|cccc}
\hline
\diagbox{Methods}{}               & G-MAE           & L-MAE           & G-RMSE           & L-RMSE          \\ \hline
non-attack          & 1.975           & -               & 4.0220           & -               \\ \hline
PGD-Random          & 5.9024          & 4.8595          & 10.3640          & 9.3635          \\
PGD-PR              & 5.7906          & 4.7277          & 10.2374          & 9.2254          \\
PGD-Centrality      & 6.0427          & 5.0215          & 10.5849          & 9.6022          \\
PGD-Degree          & 5.9408          & 4.8989          & 10.4332          & 9.4450          \\ \hline
MIM-Random          & 5.8275          & 4.7583          & 10.3654          & 9.3777          \\
MIM-PR              & 5.7358          & 4.6407          & 10.2764          & 9.2753          \\
MIM-Centrality      & 5.9866          & 4.9376          & 10.6270          & 9.6720          \\
MIM-Degree          & 5.8495          & 4.7862          & 10.3449          & 9.3616          \\ \hline
\textbf{STPGD-TDNS} & \textbf{6.1329} & \textbf{5.1647} & \textbf{10.6723} & \textbf{9.7003} \\
\textbf{STMIM-TDNS} & 5.6706          & 4.7010          & 10.1336          & 9.1813          \\ \hline
\end{tabular}
\end{adjustbox}
\end{table}
}

\subsection{Experiments at different time intervals}

We conduct further experiments at different time intervals, including 5 minutes, 10 minutes, 15 minutes, 30 minutes, and 45 minutes. We report the results at different time intervals compared with other baselines  in Tables \ref{tab:semi-PeMS-gwnet-5minutes}-\ref{tab:semi-PeMS-gwnet-45minutes}. Overall, as the time interval increases, the forecasting and adversarial attack performances decrease, as reported in Table \ref{tab:semi-PeMS-gwnet-all-time}.

For example, the G-MAE increases from 3.9458 to 6.1329 from a time interval of 5 minutes to a time interval of 60 minutes, with the attack performance degradation from 75.93\% to 67.80\%.  One possible reason is that as the time interval increases, the forecasting error of the spatiotemporal model will increase. It is more challenging for the adversarial attack methods to estimate the target label to generate effective adversarial examples.

\begin{table}[htbp]
\caption{Grey-box attack on Gwnet for \emph{PeMS-BAY} at different minutes interval }\label{tab:semi-PeMS-gwnet-all-time}
\centering
\begin{adjustbox}{width=0.9\linewidth}
\begin{tabular}{l|cccccc}
\hline
                        & 5 minutes                                               & 10 minutes                                              & 15 minutes                                              & 30 minutes                                              & 45 minutes                                              & 60 minutes                                              \\ \hline
non-attack              &  0.9496   &  1.1367  &  1.2747  &  1.6154  & 1.8872   &  1.9750   \\
STPGD-TDNS (ours)       &  3.9458   &  4.2924  &  3.6028   &  4.6629   &  5.2931   &  6.1329  \\
performance degradation &  75.93 \% &  73.46 \% &  64.62 \% &  65.36 \%&  64.34 \% &  67.80 \%\\ \hline
\end{tabular}
\end{adjustbox}
\end{table}

\begin{table}[htbp]
\caption{Grey-box attack on Gwnet for \emph{PeMS-BAY} on 5 minutes interval }\label{tab:semi-PeMS-gwnet-5minutes}
\centering
\begin{adjustbox}{width=0.6\linewidth}
\begin{tabular}{c|cccc}
\hline
                    & G-MAE           & L-MAE           & G-RMSE           & L-RMSE           \\ \hline
non-attack          & 0.9496          &                 & 1.7694           &                  \\ \hline
PGD-Random          & 3.7926          & 3.0507          & 10.1258          & 9.9924           \\
PGD-PR              & 3.8226          & 3.0885          & 10.1880          & 10.0526          \\
PGD-Centrality      & 3.7901          & 3.0586          & 10.1208          & 9.9950           \\
PGD-Degree          & 3.8302          & 3.0839          & 10.1733          & 10.0395          \\ \hline
\textbf{STPGD-TDNS} & \textbf{3.9458} & \textbf{3.2351} & \textbf{10.7429} & \textbf{10.6116} \\ \hline
\end{tabular}
\end{adjustbox}
\end{table}

\begin{table}[htbp]
\caption{Grey-box attack on Gwnet for \emph{PeMS-BAY} on 10 minutes interval }\label{tab:semi-PeMS-gwnet-10minutes}
\centering
\begin{adjustbox}{width=0.6\linewidth}
\begin{tabular}{c|cccc}
\hline
                    & \multicolumn{1}{c}{G-MAE}  & \multicolumn{1}{c}{L-MAE} & \multicolumn{1}{c}{G-RMSE} & \multicolumn{1}{c}{L-RMSE} \\ \hline
non-attack          & \multicolumn{1}{c}{1.1367} & \multicolumn{1}{c}{}      & \multicolumn{1}{c}{2.2430} & \multicolumn{1}{c}{}       \\ \hline
PGD-Random          & 4.2301                     & 3.3311                    & 10.9604                    & 10.7417                    \\
PGD-PR              & 4.2628                     & 3.3769                    & 11.0127                    & 10.7993                    \\
PGD-Centrality      & 4.2234                     & 3.3378                    & 10.9677                    & 10.7543                    \\
PGD-Degree          & 4.2779                     & 3.3778                    & 11.0364                    & 10.8219                    \\ \hline
\textbf{STPGD-TDNS} & \textbf{4.2924}            & \textbf{3.4586}           & \textbf{11.4178}           & \textbf{11.2231}           \\ \hline
\end{tabular}
\end{adjustbox}
\end{table}

\begin{table}[htbp]
\caption{Grey-box attack on Gwnet for \emph{PeMS-BAY} on 15 minutes interval }\label{tab:semi-PeMS-gwnet-15minutes}
\centering
\begin{adjustbox}{width=0.6\linewidth}
\begin{tabular}{c|cccc}
\hline
                    & G-MAE           & L-MAE           & G-RMSE          & L-RMSE          \\ \hline
non-attack          & 1.2747          &                 & 2.5761          &                 \\ \hline
PGD-Random          & 3.6073          & 2.7355          & 9.0194          & 8.6871          \\
PGD-PR              & 3.6011          & 2.7540          & 8.9609          & 8.6240          \\
PGD-Centrality      & 3.6004          & 2.7314          & 9.0132          & 8.6853          \\
PGD-Degree          & \textbf{3.6206} & 2.7510          & 8.9892          & 8.6531          \\ \hline
\textbf{STPGD-TDNS} & 3.6028          & \textbf{2.7798} & \textbf{9.1164} & \textbf{8.7607} \\ \hline
\end{tabular}
\end{adjustbox}
\end{table}

\begin{table}[htbp]
\caption{Grey-box attack on Gwnet for \emph{PeMS-BAY} on 30 minutes interval }\label{tab:semi-PeMS-gwnet-30minutes}
\centering
\begin{adjustbox}{width=0.6\linewidth}
\begin{tabular}{c|cccc}
\hline
                    & G-MAE           & L-MAE           & G-RMSE          & L-RMSE          \\ \hline
non-attack          & 1.6154          &                 & 3.2933          &                 \\ \hline
PGD-Random          & 3.4294          & 2.3903          & 6.9265          & 6.0358          \\
PGD-PR              & 3.4214          & 2.3999          & 6.8360          & 5.9331          \\
PGD-Centrality      & 3.4666          & 2.4328          & 7.0459          & 6.1729          \\
PGD-Degree          & 3.4190          & 2.3731          & 6.8721          & 5.9837          \\ \hline
\textbf{STPGD-TDNS} & \textbf{4.6629} & \textbf{3.7733} & \textbf{8.9025} & \textbf{8.1430} \\ \hline
\end{tabular}
\end{adjustbox}
\end{table}

\begin{table}[htbp]
\caption{Grey-box attack on Gwnet for \emph{PeMS-BAY} on 45 minutes interval }\label{tab:semi-PeMS-gwnet-45minutes}
\centering
\begin{adjustbox}{width=0.6\linewidth}
\begin{tabular}{c|cccc}
\hline
                    & G-MAE           & L-MAE           & G-RMSE          & L-RMSE          \\ \hline
non-attack          & 1.8872          &                 & 3.8593          &                 \\ \hline
PGD-Random          & 3.6825          & 2.4705          & 7.3557          & 6.2334          \\
PGD-PR              & 3.6789          & 2.4925          & 7.3180          & 6.1898          \\
PGD-Centrality      & 3.6872          & 2.4897          & 7.4748          & 6.3791          \\
PGD-Degree          & 3.7254          & 2.5300          & 7.4270          & 6.3325          \\ \hline
\textbf{STPGD-TDNS} & \textbf{5.2931} & \textbf{4.3660} & \textbf{9.7466} & \textbf{8.9135} \\ \hline
\end{tabular}
\end{adjustbox}
\end{table}

\end{document}